\documentclass[letterpaper]{article}

\usepackage[T1]{fontenc}
\usepackage{geometry}
\usepackage{setspace}

\usepackage{authblk}

\usepackage{booktabs}
\usepackage{amsmath,amssymb,amsthm,amsfonts}
\usepackage{xcolor}
\usepackage{array}
\usepackage{tabularx}
\usepackage{colortbl}
\usepackage{adjustbox}
\usepackage{pifont}
\usepackage{algorithm}
\usepackage{algorithmic}
\usepackage{mathptmx}
\usepackage{helvet}
\usepackage{balance}
\usepackage[hidelinks,
            colorlinks=true,
            linkcolor=blue,
            urlcolor=blue,
            citecolor=blue]{hyperref}
\usepackage{caption}
\usepackage{float}
\usepackage{textcomp}
\usepackage{stfloats}
\usepackage{verbatim}
\usepackage{placeins}
\usepackage{times}
\usepackage{moreverb}
\usepackage{mathrsfs}
\usepackage{bm}
\usepackage{xurl}
\usepackage{framed}
\usepackage[edges]{forest}
\usepackage{multicol}
\usepackage{epsfig}
\usepackage{amstext}
\usepackage{relsize}
\usepackage{lipsum}
\usepackage{longtable}
\usepackage{xtab}
\usepackage{marginnote}
\usepackage{multirow}
\usepackage{enumitem}
\usepackage{makecell}
\usepackage{wrapfig}
\usepackage{fancyhdr}
\usepackage{lastpage}
\usepackage[flushleft]{threeparttable}
\usepackage{titlesec}
\usepackage{microtype}
\usepackage{bm}

\usepackage{graphicx}
\usepackage[version=4]{mhchem} 

\usepackage[numbers,sort&compress]{natbib}
\setcitestyle{open={},close={}}

\definecolor{hidden-draw}{RGB}{20,68,106}
\definecolor{hidden-pink}{RGB}{234, 131, 121}
\definecolor{lightred}{RGB}{220,92,96}
\definecolor{deepblue}{RGB}{125,174,224}
\definecolor{lightpurp}{RGB}{179,149,189}
\definecolor{lightpurple}{RGB}{130, 132, 131}
\definecolor{lightgray}{gray}{0.9}
\definecolor{hiddenc1}{RGB}{59, 118, 122}
\definecolor{hiddenc2}{RGB}{69,105,144}
\definecolor{hiddenc3}{RGB}{130,130,170}
\definecolor{hid-vae}{RGB}{125,174,224}
\definecolor{hid-gnn}{RGB}{179,149,189}
\definecolor{hid-trans}{RGB}{122, 199,226}
\definecolor{hid-dm}{RGB}{225, 225, 255}
\definecolor{hid-llm}{RGB}{84,190,170}
\definecolor{hid-ssl}{RGB}{176,217,146}
\definecolor{hid-dms}{RGB}{238, 144, 59}
\definecolor{text-red}{RGB}{255,0,0}
\definecolor{text-blue}{RGB}{0,0,255}
\definecolor{text-green}{RGB}{0,255,0}
\definecolor{pending-yellow}{RGB}{208,181,5}
\definecolor{text-black}{RGB}{0,0,0}

\newcommand{\xs}[1]{\textcolor{text-black}{#1}}

\renewcommand{\xs}[1]{\textcolor{black}{#1}}

\newcommand{\xsrv}[1]{\textcolor{text-black}{#1}}
\newcommand{\sxsrv}[1]{\textcolor{text-black}{#1}}
\newcommand{\rv}[1]{\textcolor{text-black}{#1}}
\newcommand{\srv}[1]{\textcolor{text-black}{#1}}
\newcommand{\grv}[1]{\textcolor{text-black}{#1}}
\newcommand{\best}[1]{\textcolor{text-black}{\textbf{#1}}}
\newcommand{\second}[1]{\textcolor{text-black}{\textbf{#1}}}
\renewcommand{\best}[1]{\textbf{#1}}
\renewcommand{\second}[1]{\underline{#1}}

\newcommand{\upcite}[1]{{\color{black}(\cite{#1})}}

\titlespacing*{\section}{0pt}{8pt plus 2pt minus 2pt}{5pt plus 2pt minus 2pt}
\titlespacing*{\subsection}{0pt}{6pt plus 1pt minus 1pt}{4pt plus 1pt minus 1pt}
\titlespacing*{\paragraph}{0pt}{2pt plus 1pt minus 1pt}{1em}

\title{\srv{A Systematic Survey and Benchmark of Deep Learning for Molecular Property Prediction in the Foundation Model Era}}

\author[1,$\dagger$]{Zongru Li}
\author[1,$\dagger$]{Xingsheng Chen}
\author[1]{Honggang Wen}
\author[2,7,*]{Regina Qianru Zhang}
\author[3]{Ming Li}
\author[4]{Xiaojin Zhang}
\author[5]{Hongzhi Yin}
\author[6]{Qiang Yang}
\author[2,*]{Kwok-Yan Lam}
\author[7,*]{Pietro Lio}
\author[1,*]{Siu-Ming Yiu}

\affil[1]{The University of Hong Kong, Hong Kong SAR}
\affil[2]{Nanyang Technological University, Singapore}
\affil[3]{Zhejiang Normal University, China}
\affil[4]{The Hong Kong University of Science and Technology, Hong Kong SAR}
\affil[5]{The University of Queensland}
\affil[6]{The Hong Kong Polytechnic University, Hong Kong SAR}
\affil[7]{University of Cambridge, United Kingdom}

\date{}

\begin{document}

\maketitle

\begingroup
\renewcommand{\thefootnote}{\fnsymbol{footnote}}
\footnotetext[2]{Equal Contributions}
\footnotetext[1]{Corresponding authors; Emails: reginazhang955@gmail.com; 
kwokyan.lam@ntu.edu.sg; pl219@cam.ac.uk; smyiu@cs.hku.hk}
\endgroup

\begin{abstract}
Molecular property prediction integrates quantum chemistry, cheminformatics, and deep learning to connect molecular structure with physicochemical and biological behavior. This survey traces four complementary paradigms, including \textit{Quantum}, \textit{Descriptor Machine Learning}, \textit{Geometric Deep Learning}, and \textit{Foundation Models}, and outlines a unified taxonomy linking molecular representations, model architectures, and interdisciplinary applications. Benchmark analyses integrate evidence from both widely used datasets and datasets reflecting industry perspectives, encompassing quantum, physicochemical, physiological, and biophysical domains. The survey examines current standards in data curation, splitting strategies, and evaluation protocols, highlighting challenges including inconsistent stereochemistry, heterogeneous assay sources, and reproducibility limitations under random or poorly defined splits. These observations motivate the modernization of benchmark design toward more transparent, time- and scaffold-aware methodologies. We further propose three forward-looking directions: \grv{(i) \textit{physics-aware learning} embedding quantum consistency, (ii) \textit{uncertainty-calibrated foundation models} for trustworthy inference, and (iii) \textit{realistic multimodal benchmark ecosystems} integrating computational and experimental data.} Repository: \url{https://github.com/Zongru-Li/Survey-and-Benchmarks-of-DL-for-Molecular-Property-Prediction-in-the-Foundation-Model-Era}.
\end{abstract}

\section{Introduction}
\label{sec:intro}

Molecular property prediction~\upcite{SchNet,MPNN} stands at the confluence of quantum mechanics, chemical informatics, and artificial intelligence, aiming to map molecular structure directly to physicochemical or biological attributes. This interdisciplinary domain underlies transformative applications, from rational drug design and \rv{catalyst} development to materials discovery, yet persistent barriers in \textit{generalization}, \textit{scalability}, and \textit{uncertainty quantification} hinder universal adoption~\upcite{quantum_ml,drug_ml}. Addressing these challenges requires unifying data-driven learning with the mathematical rigor of physical chemistry while maintaining computational feasibility across vast chemical spaces.

\begin{table*}[tb]
\vspace{-0.20in}
\centering
\caption{Evolution and taxonomy of molecular property prediction approaches}\vspace{-0.05in}
\label{tab:unified_framework}
\resizebox{\linewidth}{!}{
\begin{tabular}{p{4.0cm}p{3.5cm}p{3.5cm}p{4.5cm}p{4.5cm}}
\toprule
\textbf{Historical Phase} & \textbf{Core Innovation} & \textbf{Performance Milestone} & \textbf{Prior Survey Scope} & \textbf{Advantages}\\
\midrule
\textbf{Quantum Era} \newline (1950–2000) & Wavefunction-based \newline calculations & Chemical accuracy \newline ($\sim$0.1 kcal/mol) & Theory-focused \newline (von Lilienfeld et al.~\upcite{quantum_ml}) & \textbullet~First-principles rigor \newline \textbullet~No training data required \newline \textbullet~Physically interpretable \\\midrule

\textbf{Descriptor ML} \newline (2000–2015) & Engineered molecular \newline fingerprints & Practical predictions \newline ($\sim$1.5 kcal/mol) & Pharmaceutical only \newline (Vamathevan et al.~\upcite{drug_ml}) & \textbullet~Fast inference ($\sim$ms/mol) \newline \textbullet~Low data requirements \newline \textbullet~Interpretable features \\\midrule

\textbf{Geometric Deep Learning} \newline (2015–2020) & 3D-aware graph \newline networks & Linear scaling \newline complexity & 2D structures only \newline (Wieder et al.~\upcite{gnn_review}) & \textbullet~End-to-end learning \newline \textbullet~Symmetry-aware \newline \textbullet~Scaffold generalization \\\midrule

\textbf{Foundation Models} \newline (2020–present) & Multimodal \newline pretraining & Sub-linear scaling \newline efficiency & Emerging coverage \newline (Choi et al.~\upcite{fund_model_rev}, Liyaqat et al.~\upcite{liyaqat2025advancements}) & \textbullet~Few-shot transfer \newline \textbullet~Multimodal fusion \newline \textbullet~Billion-scale corpora \\
\midrule
\multicolumn{5}{l}{\textbf{This survey's unified framework establishes:}} \\
\multicolumn{5}{l}{\quad \textbullet{} A comprehensive taxonomy linking molecular representations across dimensional hierarchies;} \\
\multicolumn{5}{l}{\quad \textbullet{} Comparative benchmarking of architectures across diverse datasets and splitting protocols;} \\
\multicolumn{5}{l}{\quad \textbullet{} A strategic synthesis of applications spanning quantum property estimation, drug discovery, and beyond.} \\
\bottomrule
\end{tabular}}
\end{table*}

Over the past seven decades, molecular property prediction has experienced a sequence of four transformative methodological revolutions (Table~\ref{tab:unified_framework}), each driven by the need to balance chemical accuracy, computational scalability, and physical interpretability. \rv{Importantly, the advent of each new paradigm did not render earlier methods obsolete. Rather, all remain actively used and continue to offer unique advantages under different contexts, ranging from first-principles rigor to large-scale generalization.} The \textbf{Quantum Era} (1950-2000) established the foundational framework for describing molecular behavior from first principles through wavefunction-based and density-functional approaches~\upcite{gomes2008calculation,unke2019physnet,gotz2013performance}. These methods deliver near-chemical accuracy and physical interpretability, \rv{key strengths that still make them indispensable for benchmarking and mechanistic insight. However, their steep computational scaling~\upcite{quantum_ml,lee1998linear,jaramillo2013large} constrains their practical use to relatively small systems. To address these limitations, the \textbf{Descriptor Machine Learning (ML) Era} (2000-2015) introduced statistical surrogates based on engineered molecular fingerprints such as ECFP~\upcite{rogers2010extended,xue2003design,ucak2023reconstruction}. These models enabled rapid inference (on the order of milliseconds per molecule) while maintaining practical predictive accuracy~\upcite{drug_ml}. Their efficiency and modest data demands remain advantageous in pharmaceutical screening and low-data regimes, though their reliance on handcrafted features limits extrapolation to unseen chemotypes. The \textbf{Geometric Deep Learning Era} (2015-2020) transformed molecular representation by incorporating 3D structural information via graph neural networks and SE(3)-equivariant architectures~\upcite{fang2022geometry,fuchs2020se}. These models combine end-to-end learning, symmetry awareness, and strong scaffold generalization, attaining accuracy levels comparable to density-functional theory~\upcite{gnn_review}. Despite these strengths, their high data and memory requirements continue to pose challenges for wide deployment. Finally, the \textbf{Foundation Model Era} (2020-present) unifies and extends prior methodologies through large-scale, multimodal pretraining~\upcite{fund_model_rev,liyaqat2025advancements,chen2023directed}. By integrating text, structure, and geometric modalities, these models achieve few-shot transfer, sub-linear scaling, and cross-domain adaptability~\upcite{wang2020generalizing,stanley2021fs}. Yet even in this new era, quantum and descriptor-based methods retain crucial roles, serve as interpretable references, efficient screening tools, or physically principled validation baselines. \textit{Together, these eras represent a cumulative, rather than replacement-based, trajectory toward unified molecular intelligence.}}

This historical trajectory underscores the central motivation of our survey: the pressing need to systematize the diverse methods of molecular AI within a coherent taxonomy (Figure~\ref{fig:taxonomy}) and unified computational pipeline (Figure~\ref{fig:pipeline}). Reviewing over one hundred deep architectures, our meta-analysis highlights three notable trends: (1) geometric GNNs tend to perform well in quantum property estimation, (2) transformer-based architectures have shown strong results in binding-affinity prediction and large-scale sequence-to-structure learning, and (3) hybrid and quantum-informed designs offer promising computational efficiencies for complex crystalline and metal–organic systems.
Collectively, these insights articulate the rationale for a new research agenda centered on \grv{(i) physics-aware learning that embeds geometric priors and quantum constraints, (ii) uncertainty-calibrated foundation models for reliable predictions in low-label and out-of-distribution settings, and (iii) developing realistic multimodal benchmark ecosystems that connect computational predictions with experimental phenomena~\upcite{alampara2025probing, cui2025multimodal, Ash2025Protocols}.}
Through this synthesis, our survey establishes not merely a literature review but also a structured framework and strategic vision for the next generation of molecular artificial intelligence. We provide a continuously updated repository \rv{at} \url{https://github.com/Zongru-Li/Survey-and-Benchmarks-of-DL-for-Molecular-Property-Prediction-in-the-Foundation-Model-Era}.

\rv{The overall structure of this paper is organized as follows: Section~\ref{sec:preliminaries} introduces the essential background concepts relevant to molecular property prediction. Section~\ref{sec:tax} presents the unified review pipeline and taxonomy that frame the comparative analysis conducted throughout the survey. Section~\ref{sec:representation} analyzes molecular representation methodologies, followed by Section~\ref{sec:architectures}, which details algorithmic architectures and compares representative deep learning approaches. Section~\ref{sec:benchmarks} presents the benchmark datasets and offers a comparative analysis of existing methods evaluated across multiple datasets, emphasizing the impact of different data splitting strategies on model performance.
Section~\ref{sec:applications} discusses practical applications across major chemical and biological domains. Finally, Section~\ref{sec:challenges_and_future} outlines the roadmap and identifies promising future research directions. The complete framework, encompassing evolution, taxonomy, benchmark evaluation, and roadmap, is illustrated in Figure~\ref{fig:intro}.}

\section{Preliminaries and Problem Definition}
\label{sec:preliminaries}
\subsection{Molecular Property Prediction Task Definition}
Molecular property prediction (MPP) refers to learning a mapping from a molecular representation to one or more properties or activities of interest~\upcite{liyaqat2025advancements}. Formally, let $m \in \mathcal{M}$ denote an input representation of a molecule (e.g., a graph, sequence, or geometric point cloud) and let $y \in \mathcal{Y}$ denote the target property. The goal is to learn a function $f:\mathcal{M}\to \mathcal{Y}$, parameterized by a deep neural network, that predicts $y$ from $m$ with high accuracy. In supervised learning settings, we assume a labeled dataset $\{(m_i, y_i)\}_{i=1}^N$ of molecules with known properties, and train $f$ by minimizing a loss function. Common prediction tasks include binary or multi-class classification\upcite{mayr_deeptox_2016, fate-tox} (e.g., active vs. inactive compounds for a target, toxicity categories) and regression of continuous properties\upcite{SchNet, ozturk_deepdta_2018} (e.g., solubility, binding affinity, or quantum mechanical energies). In regression tasks, the model predicts continuous $y\in\mathbb{R}$ and minimizes the MSE loss $\mathcal{L} = \frac{1}{N} \sum_{i=1}^N (f(m_i) - y_i)^2$. Classification tasks require the model to predict discrete labels $y \in \{1, \dots, C\}$ and minimize the cross-entropy $\mathcal{L} = -\sum_{c=1}^C y_c \log f(m)_c$. Some models~\upcite{ahmad2022chemberta,MolBERT,Wang2025Integration} address multi-task learning, predicting multiple property outputs $y = (y^{(1)}, \dots, y^{(k)})$ simultaneously for each molecule. This can improve data efficiency by sharing representations across related tasks (for instance, jointly predicting several ADMET properties). Across all formulations of molecular property prediction, a fundamental challenge lies in achieving robust generalization. The goal is to develop models that not only accurately fit known molecule-property pairs in the training data but also maintain predictive performance when applied to novel molecules~\upcite{mole,zhou2023unimol,vijil2023accelerating}. This requirement is particularly demanding because test molecules may possess structural features substantially different from those encountered during training.

The properties of interest in molecular property prediction encompass a broad spectrum across chemical, physical, and biological domains. In drug discovery applications, critical properties include pharmacokinetic and toxicological profiles, collectively referred to as ADMET properties (absorption, distribution, metabolism, excretion, and toxicity), as well as bioactivity measurements against specific disease-related molecular targets\upcite{swanson_admet-ai_2024, moleculenet, liyaqat2025advancements}. Within materials science, prediction tasks frequently involve properties such as electrical conductivity, thermal or chemical stability, and electronic band gaps of molecules and crystalline structures~\upcite{genome,bamboo,chen2025uni}. Environmental chemistry applications may focus on predicting properties including the toxicity of chemical compounds to organisms or ecosystems, as well as their biodegradability and environmental persistence~\upcite{Wang2025Integration}.  Across these diverse application domains, the prediction problem is fundamentally framed as establishing structure-property relationships. Given a molecular structure represented in an appropriate format, the task is to predict an associated label or numerical value corresponding to the property of interest.

\begin{figure*}[htb]
\vspace{-0.1in}
    \centering
    \includegraphics[width=\linewidth]{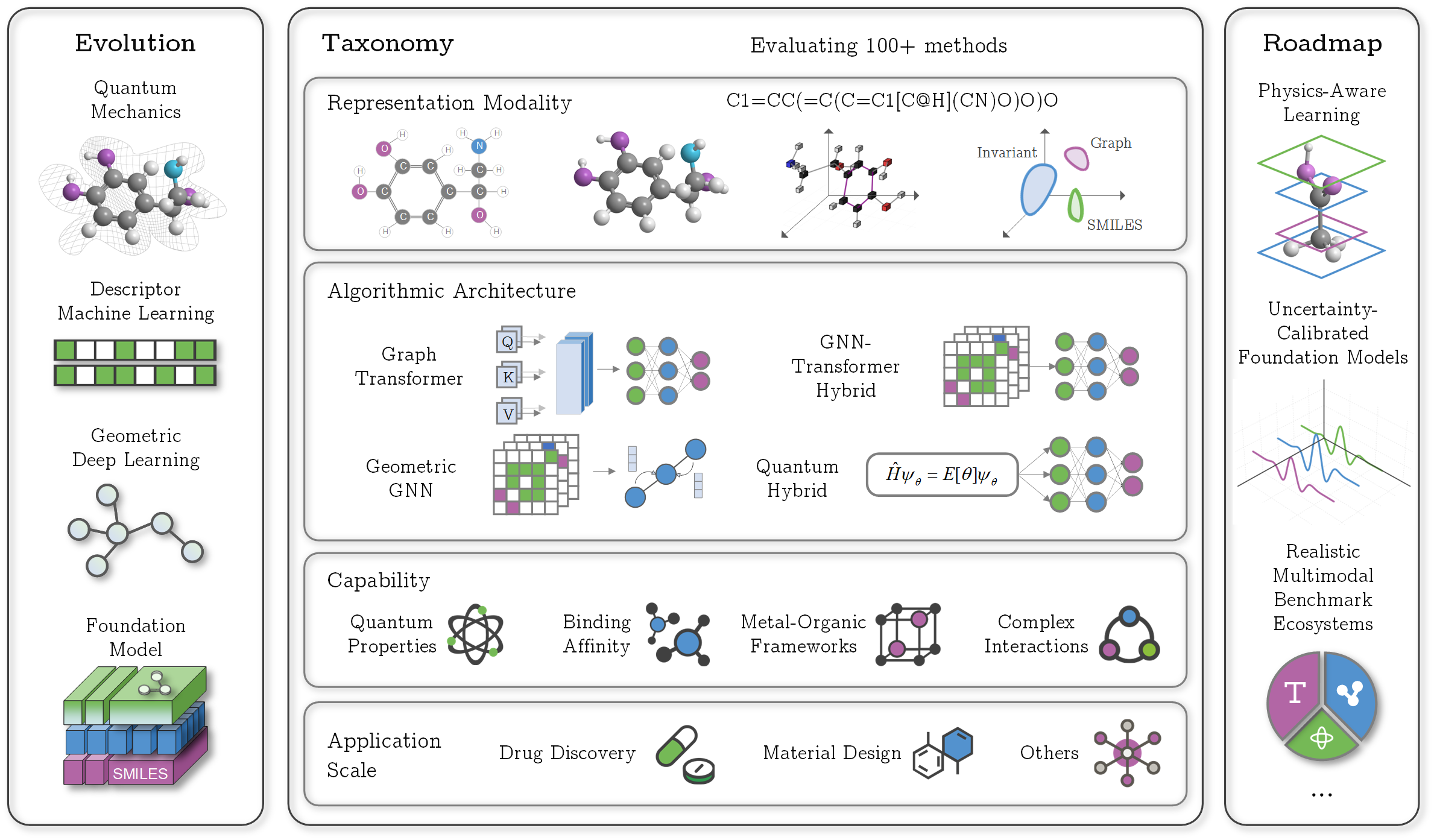}
    \caption{\rv{
\textbf{Overview of this survey.} 
The framework organizes more than 100 deep learning methods for molecular property prediction along four axes: 
\textit{Evolution}, \textit{Taxonomy}, \textit{Capability}, and \textit{Roadmap}. 
From left to right: 
(i)~\textbf{Evolution} traces the cumulative methodological trajectory from quantum mechanics and descriptor-based learning to geometric and foundation models; 
(ii)~\textbf{Taxonomy} categorizes methods by representation modality (1D–3D), algorithmic architecture (GNNs, transformers, hybrids), predictive capability (quantum, biomolecular, or materials domains), and application scale (drug discovery, materials design, etc.); 
(iii)~\textbf{Evaluation} synthesizes benchmark evidence across MoleculeNet and extended ADME datasets, using standardized data-splitting and protocol comparisons; 
(iv)~\textbf{Roadmap} distills evidence-informed and forward-looking directions: 
\textit{(a)}~\grv{physics-aware learning, supported by gains from embedding geometric priors and quantum constraints}; 
\textit{(b)}~\grv{uncertainty-calibrated foundation models, motivated by the need to distinguish reliable interpolation from risky extrapolation in low-label and out-of-distribution deployment}; 
\textit{(c)}~\grv{realistic multimodal benchmark ecosystems, derived from limitations in current dataset modalities~\upcite{alampara2025probing, cui2025multimodal} and evaluation protocols~\upcite{Ash2025Protocols}}; \grv{and other possible future directions}.
Together, these axes form a unified, evidence-based blueprint for advancing molecular AI in the foundation model era.
}}
\label{fig:intro}
\vspace{-0.20in}
\end{figure*}

\subsection{Learning Settings and Training Paradigms}
\label{sec:learning-settings}

Deep learning models for molecular prediction can be developed under various learning settings, depending on data availability and generalization requirements. Let $f_\theta$ denote a deep neural network model parameterized by $\theta$. Below we discuss each paradigm in detail. Formal mathematical definitions of the loss functions associated with these paradigms are provided in Appendix~\ref{app:losses}.

\paragraph{Supervised Learning.}
The most straightforward approach trains models on labeled datasets $\mathcal{D}_L = \{(m_i, y_i)\}_{i=1}^N$ of $N$ molecules with ground-truth properties. Given sufficient labeled examples, supervised deep networks (GNNs, CNNs, Transformers) can achieve excellent performance by minimizing a task-specific loss $\mathcal{L}_{\mathrm{task}}(f_\theta(m_i), y_i)$ over all training pairs~\upcite{MPNN,SchNet,Graphormer}. However, data scarcity ($N$ is small) poses a significant challenge, as obtaining experimental labels is time-consuming and expensive. Models also struggle with out-of-distribution (OOD) molecules not covered by the training distribution $P_{\mathrm{train}}(\mathcal{M}, \mathcal{Y})$, making purely supervised approaches insufficient for exploring the full breadth of chemical space~\upcite{mole,zhou2023unimol,vijil2023accelerating}.

\paragraph{Self-Supervised Learning and Pretraining.}
To leverage abundant unlabeled chemical data $\mathcal{D}_U = \{m_i\}_{i=1}^M$ (typically $M \gg N$), self-supervised learning (SSL) has emerged as a powerful paradigm~\upcite{graphmae,chemberta,MolBERT}. Models are pretrained on $\mathcal{D}_U$ through proxy tasks that require no human-provided labels, encouraging the model to learn meaningful molecular representations $z = f_\theta(m)$ by minimizing a self-supervised loss $\mathcal{L}_{\mathrm{SSL}}$. Common strategies include:
\begin{itemize}
    \item \emph{Contrastive learning} (e.g., MolCLR~\upcite{wang2022molecular}): maximizes the agreement (e.g., cosine similarity) between representations of different augmentations (views) $m'$ and $m''$ of the same molecule, while pushing representations of different molecules apart.
    \item \emph{Masking tasks} (e.g., GraphMAE~\upcite{graphmae}): corrupts the input molecule $m$ to create $\tilde{m}$ (e.g., by masking node features) and trains the model to reconstruct the original information, minimizing a reconstruction loss $\mathcal{L}_{\mathrm{recon}}(f_\theta(\tilde{m}), m)$.
\end{itemize}
Pretraining on millions of molecules can significantly boost downstream performance. Large-scale pretraining is now feasible: the MolE model~\upcite{mole}, pretrained on $\mathcal{D}_U$ with $M \approx 842$ million molecules via self-supervised and multi-task objectives, achieved state-of-the-art results. These chemistry foundation models~\upcite{zhou2023unimol,genome,vijil2023accelerating,huang_foundation_2024,nomura2025allegro} aim to learn universal chemical representations that transfer across tasks.

\paragraph{Transfer Learning and Fine-Tuning.}
Transfer learning adapts models pretrained on one dataset to target tasks. The typical approach involves obtaining pretrained parameters $\theta_{\mathrm{pre}}$ (often via SSL on $\mathcal{D}_U$ or multi-task learning), then performing supervised fine-tuning on a smaller, task-specific dataset $\mathcal{D}_L$ starting from $\theta = \theta_{\mathrm{pre}}$. For example, Transformer models~\upcite{smilesbert, chemberta, MolBERT, daTransformer} pretrained on unlabeled SMILES can be fine-tuned to predict properties like solubility, often outperforming models trained from scratch. Transfer learning also occurs through multi-task learning, which trains on $K$ related tasks simultaneously by minimizing a weighted combination $\sum_{k=1}^K \lambda_k \mathcal{L}_{\mathrm{task}}^{(k)}$, or through few-shot learning techniques using meta-learning~\upcite{Wang2025Integration}.

\paragraph{Unsupervised and Semi-Supervised Settings.}
Some scenarios integrate unlabeled ($\mathcal{D}_U$) and labeled ($\mathcal{D}_L$) data in semi-supervised approaches. This often involves a combined loss, $\mathcal{L}_{\mathrm{total}} = \mathcal{L}_{\mathrm{supervised}} + \lambda\, \mathcal{L}_{\mathrm{unsupervised}}$, allowing the model to propagate label information from $\mathcal{D}_L$ using the underlying structure learned from $\mathcal{D}_U$. Unsupervised evaluation of molecular embeddings $z = f_\theta(m)$ ensures learned representations capture meaningful chemistry, such as clustering molecules by functional class in the latent space~\upcite{mol2vec}. 

\paragraph{Practical considerations.}
The choice among these paradigms hinges on several practical factors. When labeled data exceeds several thousand examples per task and the target domain overlaps with training chemistry, supervised learning often suffices. SSL pretraining provides the largest gains when downstream labels are scarce (typically below a few hundred), the target property domain differs substantially from pretraining data, and the unlabeled corpus is chemically diverse~\upcite{mole,chemberta}. Multi-task transfer is most effective when endpoint properties are biochemically related (e.g., multiple ADMET assays), allowing shared representations to compensate for per-task label sparsity. Semi-supervised approaches occupy a middle ground suited to settings where moderate labeled and abundant unlabeled data coexist within a consistent chemical domain.

\subsection{Data Splitting Methods}
\label{sec:data_split_definition}
\rv{Several dataset splitting strategies have been proposed for evaluating molecular property prediction models, each providing a different degree of separation between training and test sets.
The most straightforward approach, \textit{random splitting}, partitions the data randomly into training and test sets at a fixed ratio (typically 80/20). When hyperparameter tuning is required, a validation set is additionally held out from the training portion. A common refinement is \textit{stratified random splitting}, which preserves the distribution of the target variable (e.g., active/inactive ratio for classification, or value range for regression) across all partitions, reducing the risk of sampling bias in imbalanced datasets.
\textit{Scaffold splitting}, popularized by the MoleculeNet benchmark~\upcite{moleculenet}, leverages the Bemis--Murcko decomposition~\upcite{bemis1996properties}, which iteratively removes all side-chain atoms from a molecule until only the core ring systems and their connecting linkers remain. Each molecule is then assigned a scaffold label, and the splitting procedure ensures that all molecules sharing the same scaffold are assigned exclusively to either the training or test set.
To achieve greater structural separation without relying on predefined substructure rules, \textit{fingerprint-based clustering} methods have been employed. A representative example is \textit{Butina clustering}~\upcite{butina1999unsupervised}, which operates on molecular fingerprints (typically Morgan/ECFP fingerprints) and groups molecules whose pairwise Tanimoto similarity exceeds a predefined threshold. The algorithm first identifies cluster centers as molecules with the greatest number of neighbors within the threshold and then assigns remaining molecules to the nearest center. Molecules within the same cluster are then kept together in either the training or test set.
More recently, Guo et al.~\upcite{guo2024scaffold,guo2025umap} proposed a \textit{UMAP-based clustering split}. In this approach, Morgan fingerprints are first projected into a low-dimensional (typically two-dimensional) space via Uniform Manifold Approximation and Projection (UMAP), a nonlinear dimensionality reduction method that preserves local neighborhood structure while often retaining useful global organization for clustering. Agglomerative clustering is then applied to the projected coordinates to partition molecules into $k$ groups, each of which is assigned entirely to the training or test set.}

\rv{Finally, \textit{time-based (temporal) splits} partition the dataset according to the chronological order in which compounds were assayed~\upcite{sheridan2013timesplit}. Models are trained on historically earlier data and evaluated on later compounds, directly mimicking the prospective deployment scenario in which predictions are made for newly synthesized molecules. For example, given a year of assay data, one may train on compounds tested between January and September and evaluate on those tested between October and December~\upcite{walters2024splitting}.}

\subsection{Uncertainty Quantification}
\label{sec:uq}
 
\srv{When a molecular property prediction model outputs a point estimate $\hat{y} = f_\theta(m)$, practitioners also assess how much confidence to place in that prediction. Uncertainty quantification (UQ) assigns a measure of reliability to each prediction, enabling downstream decisions such as whether to trust a virtual screening hit, which compounds to prioritize for experimental validation, and how to allocate resources in active learning loops. Given the high cost of wet-lab experiments in drug discovery and materials design, reliable UQ is essential for translating computational predictions into actionable scientific outcomes. This issue becomes particularly acute in the foundation-model regime, where pretrained representations are deployed precisely for low-label transfer and out-of-distribution generalization, making calibrated uncertainty a prerequisite for trustworthy use.}
 
\srv{\paragraph{Sources of uncertainty.} Prediction uncertainty is conventionally decomposed into two complementary components~\upcite{dai2025uncertainty, nigam2021assigning}. \textit{Aleatoric} (data) uncertainty arises from inherent noise or ambiguity in the training labels. For example, binding affinity measurements conducted under heterogeneous assay conditions, or the use of approximate exchange-correlation functionals in DFT-derived labels. This form of uncertainty is irreducible even with infinite data. \textit{Epistemic} (model) uncertainty reflects limitations in the model itself: insufficient training data, restricted model capacity, or poor coverage of the input space. Unlike aleatoric uncertainty, epistemic uncertainty can in principle be reduced by acquiring more informative data or improving the model.}

\srv{\paragraph{UQ methodologies.}
Following the categorization of Hirschfeld et al.~\upcite{hirschfeld2020uncertainty}, we organize the primary UQ strategies for neural network-based molecular property prediction into four families: ensemble-based, mean-variance estimation, distance-based, and union-based methods. We also briefly note additional methods beyond this taxonomy. For broader coverage of the UQ landscape, we refer readers to dedicated reviews and comparative benchmark studies~\upcite{dai2025uncertainty, nigam2021assigning, hirschfeld2020uncertainty}.}

\srv{\textit{Ensemble-based methods} train multiple models and use the variance of their predictions as an uncertainty estimate. Diversity among ensemble members can be introduced through different parameter initializations (traditional ensembling), training on different data subsets (bootstrapping), periodically saving model snapshots during training (snapshot ensembling), or applying stochastic dropout masks at inference time (Monte Carlo dropout)~\upcite{hirschfeld2020uncertainty, scalia2020evaluating}. Ensembles are conceptually straightforward, can be parallelized, and have been shown in benchmark studies to provide a strong practical UQ baseline across molecular datasets~\upcite{dai2025uncertainty, hirschfeld2020uncertainty}. However, their computational cost scales linearly with ensemble size, which can be prohibitive for large-scale screening.}

\srv{\textit{Mean-variance estimation} (MVE) modifies a single neural network to jointly predict a mean $\mu(x)$ and variance $\sigma^2(x)$ for each input, trained by minimizing a Gaussian negative log-likelihood loss. Because both outputs are produced in a single forward pass, MVE offers substantial efficiency advantages over ensembles. However, the predicted variance may be miscalibrated without additional post-hoc adjustment~\upcite{dai2025uncertainty, hirschfeld2020uncertainty}.}

\srv{\textit{Distance-based methods} estimate uncertainty based on how far a test molecule lies from the training data in a molecular representation space. In structure space, fingerprint dissimilarity measures such as Tanimoto distance are commonly used. In latent space, Euclidean distance in a learned embedding can reflect task-relevant distributional shift~\upcite{hirschfeld2020uncertainty}. Related approaches include nearest-neighbor error estimation such as loss trajectory analysis for uncertainty (LTAU)~\upcite{vita2025ltau} and density-based methods using Gaussian mixture models in feature space~\upcite{zhu2023fast}. Distance-based methods are computationally efficient at inference, but they quantify distributional shift rather than prediction error directly and typically require recalibration to serve as quantitative error estimates~\upcite{dai2025uncertainty}.}

\srv{\textit{Union-based methods} combine the representational power of neural networks with secondary models that natively produce uncertainty estimates~\upcite{hirschfeld2020uncertainty}. In this pipelined approach, a neural network is first trained for the regression task, after which its output layer is removed to obtain an embedding network, and the resulting feature vectors are used to train a secondary model, such as a Gaussian process or random forest, that produces both a prediction and a quantitative uncertainty estimate. Hirschfeld et al.~\upcite{hirschfeld2020uncertainty} found that a message passing network coupled with a random forest (MPNN RF) performed best overall across their benchmark, making union-based methods a strong candidate in that study.}

\srv{Beyond these four families, several other UQ paradigms have been applied to molecular property prediction. \textit{Evidential regression} places a higher-order prior (e.g., Normal Inverse-Gamma) over the likelihood parameters and trains the network to output the evidential hyperparameters, yielding both prediction and uncertainty from a single forward pass~\upcite{amini2020deep, soleimany2021evidential}. \textit{Conformal prediction} can provide prediction intervals with formal finite-sample coverage guarantees under exchangeability assumptions~\upcite{nigam2021assigning}. \textit{Bayesian neural networks} place distributions over network weights and approximate the intractable posterior via variational inference or sampling. Monte Carlo dropout can also be interpreted as a variational Bayesian approximation~\upcite{gal2016dropout}, bridging Bayesian and ensemble perspectives. \textit{Gaussian processes} provide exact Bayesian inference for small datasets but scale cubically with dataset size, limiting their standalone use for large molecular libraries~\upcite{liu2020gaussian, nigam2021assigning}.}

\tikzstyle{my-box}=[
    rectangle,
    draw=hidden-draw,
    rounded corners,
    align=center,
    text opacity=1,
    minimum height=1.5em,
    minimum width=5em,
    inner sep=2pt,
    fill opacity=.8,
    line width=0.8pt,
]

\tikzstyle{leaf-head}=[my-box, minimum height=1.5em,
    draw=black, 
    text=black, font=\normalsize,
    inner xsep=2pt,
    inner ysep=4pt,
    line width=0.8pt,
]

\tikzstyle{leaf-task}=[my-box, minimum height=2.5em,
    draw=black, 
    align=left,
    text=black, font=\normalsize,
    inner xsep=2pt,
    inner ysep=4pt,
    line width=0.8pt,
]
\tikzstyle{leaf-taska}=[my-box, minimum height=2.5em,
    draw=black, 
    align=left,
    text=black, font=\normalsize,
    inner xsep=2pt,
    inner ysep=4pt,
    line width=0.8pt,
]
\tikzstyle{leaf-task1}=[my-box, minimum height=2.5em,
    draw=black, 
    text=black, font=\normalsize,
    inner xsep=2pt,
    inner ysep=4pt,
    line width=0.8pt,
]
\tikzstyle{modelnode-task1}=[my-box, minimum height=1.5em,
    draw=black, 
    text=black, font=\normalsize,
    inner xsep=2pt,
    inner ysep=4pt,
    line width=0.8pt,
]
\tikzstyle{leaf-task2}=[my-box, minimum height=2.5em,
    draw=black, 
    text=black, font=\normalsize,
    inner xsep=2pt,
    inner ysep=4pt,
    line width=0.8pt,
]
\tikzstyle{modelnode-task2}=[my-box, minimum height=1.5em,
    draw=black, 
    text=black, font=\normalsize,
    inner xsep=2pt,
    inner ysep=4pt,
    line width=0.8pt,
]
\tikzstyle{leaf-task3}=[my-box, minimum height=2.5em,
    draw=black, 
    text=black, font=\normalsize,
    inner xsep=2pt,
    inner ysep=4pt,
    line width=0.8pt,
]
\tikzstyle{modelnode-task3}=[my-box, minimum height=1.5em,
    draw=black, 
    text=black, font=\normalsize,
    inner xsep=2pt,
    inner ysep=4pt,
    line width=0.8pt,
]

\tikzstyle{leaf-task4}=[my-box, minimum height=2.5em,
    draw=black, 
    text=black, font=\normalsize,
    inner xsep=2pt,
    inner ysep=4pt,
    line width=0.8pt,
]
\tikzstyle{modelnode-task4}=[my-box, minimum height=1.5em,
    draw=black, 
    text=black, font=\normalsize,
    inner xsep=2pt,
    inner ysep=4pt,
    line width=0.8pt,
]

\tikzstyle{leaf-task5}=[my-box, minimum height=2.5em,
    draw=black, 
    text=black, font=\normalsize,
    inner xsep=2pt,
    inner ysep=4pt,
    line width=0.8pt,
]

\tikzstyle{modelnode-task5}=[my-box, minimum height=1.5em,
    draw=black, 
    text=black, font=\normalsize,
    inner xsep=2pt,
    inner ysep=4pt,
    line width=0.8pt,
]

\tikzstyle{leaf-task6}=[my-box, minimum height=2.5em,
    draw=black, 
    text=black, font=\normalsize,
    inner xsep=2pt,
    inner ysep=4pt,
    line width=0.8pt,
]

\tikzstyle{leaf-task7}=[my-box, minimum height=2.5em,
    draw=black, 
    text=black, font=\normalsize,
    inner xsep=2pt,
    inner ysep=4pt,
    line width=0.8pt,
]
\tikzstyle{leaf-task8}=[my-box, minimum height=2.5em,
    draw=black, 
    text=black, font=\normalsize,
    inner xsep=2pt,
    inner ysep=4pt,
    line width=0.8pt,
]

\tikzstyle{leaf-task9}=[my-box, minimum height=2.5em,
    draw=black, 
    text=black, font=\normalsize,
    inner xsep=2pt,
    inner ysep=4pt,
    line width=0.8pt,
]
\tikzstyle{leaf-task10}=[my-box, minimum height=2.5em,
    draw=black, 
    align=left,
    text=black, font=\normalsize,
    inner xsep=2pt,
    inner ysep=4pt,
    line width=0.8pt,
]

\tikzstyle{modelnode-task6}=[my-box, minimum height=1.5em,
    draw=black, 
    text=black, font=\normalsize,
    inner xsep=2pt,
    inner ysep=4pt,
    line width=0.8pt,
]

\tikzstyle{modelnode-task7}=[my-box, minimum height=1.5em,
    draw=black, 
    text=black, font=\normalsize,
    inner xsep=2pt,
    inner ysep=4pt,
    line width=0.8pt,
]
\tikzstyle{modelnode-task8}=[my-box, minimum height=1.5em,
    draw=black, 
    text=black, font=\normalsize,
    inner xsep=2pt,
    inner ysep=4pt,
    line width=0.8pt,
]
\tikzstyle{modelnode-task9}=[my-box, minimum height=1.5em,
    draw=black, 
    text=black, font=\normalsize,
    inner xsep=2pt,
    inner ysep=4pt,
    line width=0.8pt,
]

\tikzstyle{leaf-paradigms}=[
    my-box,
    minimum height=2.5em,
    align=left,
    draw=black, 
    text=black, font=\normalsize,
    inner xsep=2pt,
    inner ysep=4pt,
    line width=0.8pt,
]
\tikzstyle{leaf-others}=[my-box, minimum height=2.5em,
    draw=black, 
    text=black, font=\normalsize,
    inner xsep=2pt,
    inner ysep=4pt,
    line width=0.8pt,
]
\tikzstyle{leaf-other}=[my-box, minimum height=2.5em,
    draw=orange!80, 
    fill=orange!15,  
    text=black, font=\normalsize,
    inner xsep=2pt,
    inner ysep=4pt,
    line width=0.8pt,
]

\tikzstyle{modelnode-task}=[my-box, minimum height=1.5em,
    draw=black, 
    text=black, font=\normalsize,
    inner xsep=2pt,
    inner ysep=4pt,
    line width=0.8pt,
]

\tikzstyle{modelnode-paradigms}=[my-box, minimum height=1.5em,
    draw=black, 
    text=black, font=\normalsize,
    inner xsep=2pt,
    inner ysep=4pt,
    line width=0.8pt,
]
\tikzstyle{modelnode-others}=[my-box, minimum height=1.5em,
    draw=black, 
    text=black, font=\normalsize,
    inner xsep=2pt,
    inner ysep=4pt,
    line width=0.8pt,
]
\tikzstyle{modelnode-other}=[my-box, minimum height=1.5em,
    draw=black, 
    text=black, font=\normalsize,
    inner xsep=2pt,
    inner ysep=4pt,
    line width=0.8pt,
]
\begin{figure*}[!ht]
    \centering
    \resizebox{1\textwidth}{!}{
        \begin{forest}
            forked edges,
            for tree={
                grow=east,
                reversed=true,
                anchor=base west,
                parent anchor=east,
                child anchor=west,
                base=left,
                font=\normalsize,
                rectangle,
                draw=hidden-draw,
                rounded corners,
                align=left,
                minimum width=1em,
                edge+={darkgray, line width=1pt},
                s sep=3pt,
                inner xsep=0pt,
                inner ysep=3pt,
                line width=0.8pt,
                ver/.style={rotate=90, child anchor=north, parent anchor=south, anchor=center},
            }, 
            [
                Deep Learning for Molecular Property Prediction,leaf-head, ver
                [
                    Representation\\ Modalities\\(\textbf{Sec.~\ref{sec:representation}}), leaf-paradigms,text width=8em
                    [
                        1D Representations\\(\textbf{Sec.~\ref{sec:1d-rep}}), leaf-paradigms, text width=9.5em
                        [
                            SMILES\\(\textbf{Sec.~\ref{sec:smiles}}), leaf-paradigms, text width=7.5em
                            [
                                {{ }SimSon~\upcite{simson}, ChemBERTa~\upcite{chemberta}, MolBERT~\upcite{MolBERT}, daTransformer~\upcite{daTransformer}, SCFP~\upcite{scfp},\\
                                { }DeepSMILES~\upcite{deepsmiles}, Li \textit{et al.}~\upcite{spe}, Smiles-BERT~\upcite{smilesbert}, SPVec~\upcite{spvec}, Mol2vec~\upcite{mol2vec},\\
                                { }Alhmoudi \textit{et al.}~\upcite{Alhmoudi2025}, DeepDTA~\upcite{ozturk_deepdta_2018}, ChemBERTa-2~\upcite{ahmad2022chemberta}, Mol-BERT~\upcite{li2021mol},\\
                                { }Chemformer~\upcite{chemformer}, Shin \textit{et al.}~\upcite{shin_self-attention_2019}, SMILES2vec~\upcite{smile2vec}, ReactionT5~\upcite{reactiont5},\\
                                { }Born \textit{et al.}~\upcite{born2023chemical}, MolTrans~\upcite{huang_moltrans_2021}},
                                text width = 34.25em
                            ]
                        ]
                        [
                            SELFIES\\(\textbf{Sec.~\ref{sec:selfies}}), leaf-paradigms, text width=7.5em
                            [{{ }Alhmoudi \textit{et al.}~\upcite{Alhmoudi2025}, Group SELFIES~\upcite{groupselfies}, SELFormer~\upcite{selformer}, SELFIES~\upcite{selfies}}, text width = 34.25em]
                        ]
                        [
                            Other Sequences\\(\textbf{Sec.~\ref{sec:sequence}}), leaf-paradigms, text width=7.5em
                            [{{ }word2vec~\upcite{word2vec}, InChI~\upcite{inchi}, Mayr \textit{et al.}~\upcite{mayr_deeptox_2016}}, text width = 34.25em]
                        ]   
                    ]
                    [
                        Molecule Topological\\Graph (2D)\\(\textbf{Sec.~\ref{sec:2d-rep}}), leaf-paradigms, text width=9.5em
                        [
                            {{ }GAT~\upcite{velickovic2018graph}, Graphormer~\upcite{Graphormer}, MolE~\upcite{mole}, MolCLR~\upcite{wang2022molecular}, GNN-SKAN~\upcite{li2025gnn}, GCN~\upcite{kipf2017semi}, SPECTRA~\upcite{nogueira2025spectraspectraltargetawaregraph},\\
                            { }Rampasek \textit{et al.}~\upcite{rampavsek2022recipe}, GraphMAE~\upcite{graphmae}, Wieder \textit{et al.}~\upcite{gnn_review}, N-Gram Graph~\upcite{liu2019n}, Gilmer \textit{et al.}~\upcite{MPNN},\\
                            { }Sato \textit{et al.}~\upcite{sato2024chemical}, Rong \textit{et al.}~\upcite{rong2020self}, Xu \textit{et al.}~\upcite{xu2018powerful}, Bresson \textit{et al.}~\upcite{bresson2024kagnns},
                            { }GraphKAN~\upcite{ahmed2024graphkan}, D-MPNN~\upcite{MPNN}},
                            text width = 43.25em
                        ]       
                    ]
                    [
                        Geometric \\Conformation (3D)\\(\textbf{Sec.~\ref{sec:3d-rep}}), leaf-paradigms, text width=9.5em
                        [
                            {{ }Lilienfeld \textit{et al.}~\upcite{quantum_ml}, DimeNet~\upcite{dimenet}, SchNet~\upcite{SchNet}, TorchMD-Net~\upcite{tholke2022torchmd}, SphereNet~\upcite{spherenet}, Uni-Mol+~\upcite{lu2023highly},\\
                            { }Gasteiger \textit{et al.}~\upcite{gasteiger2020fast}, Thomas \textit{et al.}~\upcite{thomas2018tensor}, DiffDock~\upcite{diffdock}, Shi \textit{et al.}~\upcite{shi2022benchmarking}, GemNet~\upcite{gasteiger2021gemnet},\\
                            { }GPS++~\upcite{masters2022gps++}, EGNN~\upcite{EGNN}, Allegro-FM~\upcite{nomura2025allegro}, Yuan \textit{et al.}~\upcite{yuan2023molecular},
                            { }Maziarka \textit{et al.}~\upcite{maziarka2020molecule}, \\{ }ChemRL-GEM~\upcite{fang2022geometry}},
                            text width = 43.25em
                        ]   
                    ]
                    [
                        Multimodal\\(\textbf{Sec.~\ref{sec:mult-rep}}), leaf-paradigms, text width=9.5em
                        [
                            { }Liu \textit{et al.}~\upcite{liu2021pre}{, } Graph-BERT~\upcite{jha2023graph}{, } Holo-Mol~\upcite{mao2024holo}{, } GraSeq~\upcite{guo2020graseq}\\, text width = 43.25em
                        ]   
                    ]
                ]
                [
                     Model Architecture\\(\textbf{Sec.~\ref{sec:architectures}}), leaf-task,text width=8em
                    [
                        Geometric GNNs\\(\textbf{Sec.~\ref{sec:gnn}}), leaf-task10, text width=9.5em
                        [
                            { }SchNet~\upcite{SchNet}{, } EGNN~\upcite{EGNN}{, } SE(3)-transformers~\upcite{fuchs2020se}{, } Allegro-FM~\upcite{nomura2025allegro}{, } SphereNet~\upcite{spherenet}{, }\\{ }ChemRL-GEM~\upcite{fang2022geometry}{, } GemNet~\upcite{gasteiger2021gemnet}{, } Yuan \textit{et al.}~\upcite{yuan2023molecular}{, } Gasteiger \textit{et al.}~\upcite{gasteiger2020fast}{, } DiffDock~\upcite{diffdock}{, }\\{ }Tensor field networks~\upcite{thomas2018tensor}, text width=43.25em
                        ]
                    ]
                    [
                        Graph Transformer\\(\textbf{Sec.~\ref{sec:transformer}}), leaf-taska, text width=9.5em
                            [
                                { }MolE~\upcite{mole}{, }
                                SimSon~\upcite{simson}{, } Graphormer~\upcite{Graphormer}{, } Shi \textit{et al.}~\upcite{shi2022benchmarking}{, }Yuan \textit{et al.}~\upcite{yuan2023molecular}{, } Torchmd-net~\upcite{tholke2022torchmd}{, } \\ { }Maziarka \textit{et al.}~\upcite{maziarka2020molecule}{, } Sato \textit{et al.}~\upcite{sato2024chemical}{, } Rampasek \textit{et al.}~\upcite{rampavsek2022recipe}{, } Chen \textit{et al.}~\upcite{chen2023directed}{, } Rong \textit{et al.}~\upcite{rong2020self} \\, text width=43.25em
                            ]
                    ]
                    [
                        Hybrid Architecture\\(\textbf{Sec.~\ref{sec:hybrid}}), leaf-taska, text width=9.5em
                            [   
                                {{ }Chemception~\upcite{chemception}, GPS++~\upcite{masters2022gps++}, Pre-training with 3D geometry~\upcite{liu2021pre}, GraSeq~\upcite{guo2020graseq}, KAGNNs~\upcite{bresson2024kagnns},\\
                                { }GNN-SKAN~\upcite{li2025gnn}, Graph-BERT~\upcite{jha2023graph}, GraphKAN~\upcite{ahmed2024graphkan}, Holo-Mol~\upcite{mao2024holo}},
                                text width=43.25em
                            ]
                    ]
                    [
                        Quantum Hybrid\\(\textbf{Sec.~\ref{sec:quantum}}), leaf-taska, text width=9.5em
                            [
                            {{ }PennyLane~\upcite{arrazola2021differentiable}, Gao \textit{et al.}~\upcite{gao2023generalizing}},
                            text width=43.25em
                            ]
                    ]
                ]
                [
                     Applications\\(\textbf{Sec.~\ref{sec:applications}}), leaf-task,text width=8em
                    [
                        Drug Discovery\\(\textbf{Sec.~\ref{sec:drug}}), leaf-taska, text width=9.5em
                            [
                                {{ }ChemBERTa~\upcite{chemberta}, MolBERT~\upcite{MolBERT}, ChemBERTa-2~\upcite{ahmad2022chemberta}, DiffDock~\upcite{diffdock}, EGNN~\upcite{EGNN}, GraphDTA~\upcite{nguyen_graphdta_2021},\\
                                { }SELFormer~\upcite{selformer}, MolE~\upcite{mole}, Gilmer \textit{et al.}~\upcite{MPNN}, Alhmoudi \textit{et al.}~\upcite{Alhmoudi2025}, MolTrans~\upcite{huang_moltrans_2021}, ADMET-AI~\upcite{swanson_admet-ai_2024},\\
                                { }Gasteiger \textit{et al.}~\upcite{gasteiger2020fast}, Yuan \textit{et al.}~\upcite{yuan2023molecular}, Rong \textit{et al.}~\upcite{rong2020self}, SPECTRA~\upcite{nogueira2025spectraspectraltargetawaregraph},
                                { }DeepDTA~\upcite{ozturk_deepdta_2018}, GAT~\upcite{velickovic2018graph}, \\ { }Chemprop~\upcite{heid_chemprop_2024}, FATE-Tox~\upcite{fate-tox}, ChemRL-GEM~\upcite{fang2022geometry},
                                { }Chemception~\upcite{chemception}, MolCLR~\upcite{wang2022molecular}},
                                text width=43.25em
                            ]
                    ]
                    [
                        Materials Design\\(\textbf{Sec.~\ref{sec:material}}), leaf-taska, text width=9.5em
                            [
                                {{ }Lilienfeld \textit{et al.}~\upcite{quantum_ml}, SchNet~\upcite{SchNet}, BAMBOO~\upcite{bamboo}, Gao \textit{et al.}~\upcite{gao2023generalizing}, Razakh \textit{et al.}~\upcite{razakh2021pnd}, GNoME~\upcite{genome},\\
                                { }CatBERTa~\upcite{ock2023catberta}, CataLM~\upcite{catalm}, Uni-Electrolyte~\upcite{chen2025uni}, Allegro-FM~\upcite{nomura2025allegro}, CatGPT~\upcite{catgpt}},
                                text width=43.25em
                            ]
                    ]
                    [
                        Others\\(\textbf{Sec.~\ref{sec:otherapp}}), leaf-taska, text width=9.5em
                            [
                                {{ }Thomas~\upcite{thomas2018tensor}, Shin \textit{et al.}~\upcite{shin_self-attention_2019}, ChemFormer~\upcite{chemformer}, Molecular Transformer~\upcite{schwaller2019molecular}, Wang \textit{et al.}~\upcite{Wang2025Integration}},
                                text width=43.25em
                            ]
                    ]
                ]
            ]
        \end{forest}
        }
    \caption{Taxonomy of existing studies for molecular property prediction}
    \label{fig:taxonomy}
\end{figure*}
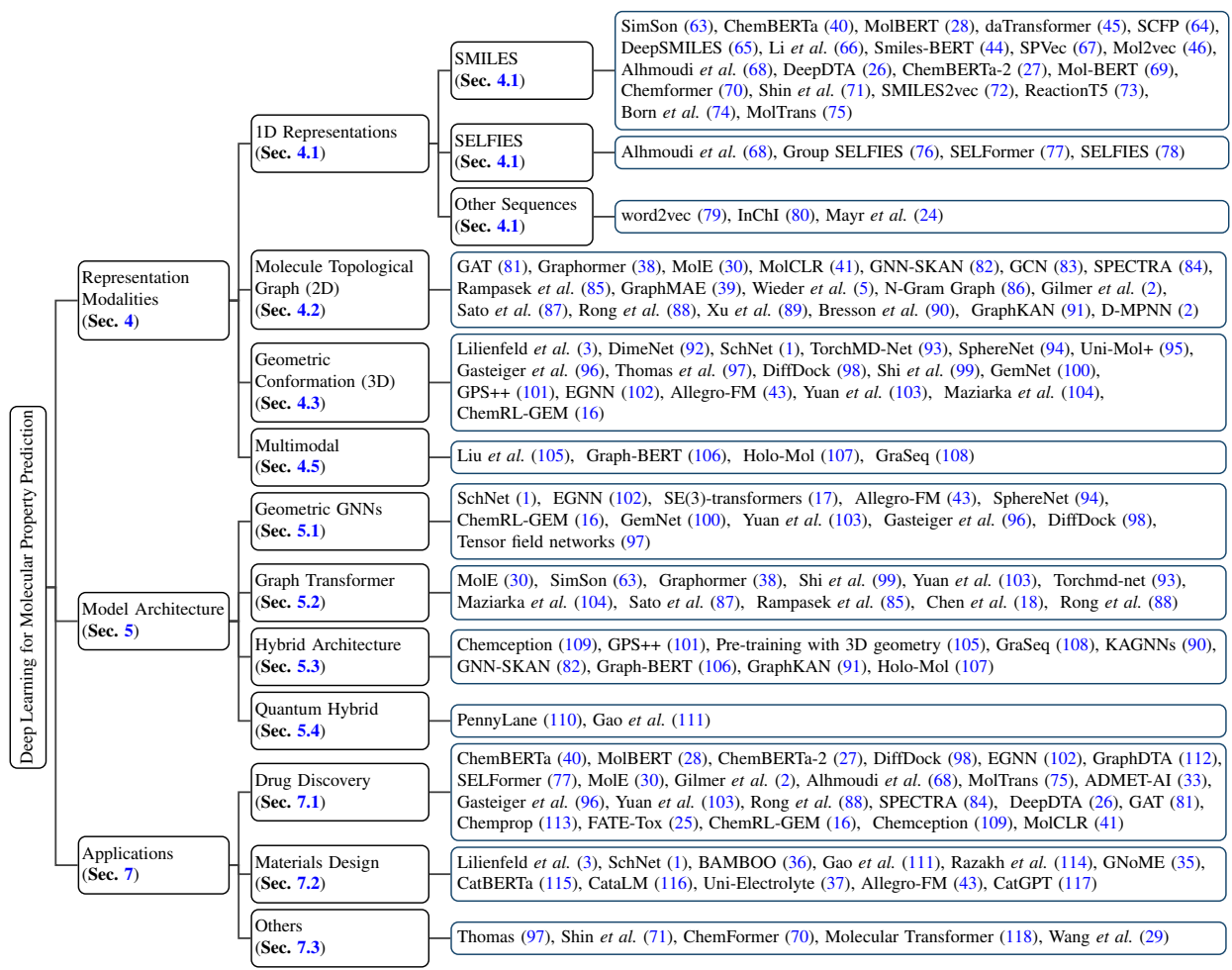

\section{Pipeline and Taxonomy} 
\label{sec:tax}

\rv{This survey provides a unified framework that organizes prevailing MPP methodologies~\upcite{gnn_review,liyaqat2025advancements} across representation modality, algorithmic architecture, and fields of application. By integrating critical deep learning approaches into three MPP-specific aspects, our synthesis facilitates cross-method comparison and identifies underexplored synergies among methods of varying attributes. First, it delineates a generalized pipeline representative of best practices in MPP, starting with datasets of diverse molecular representations, proceeding through deep learning architectures and optimization strategies to distill meaningful patterns, and culminating in predictive models for analysis tasks. Second, the survey introduces a comprehensive taxonomy that classifies relevant deep learning implementations based on input modality, models' inherent architecture, and application scenarios.}

\subsection{Pipeline}

\begin{figure*}
    \centering
    \includegraphics[width=\linewidth]{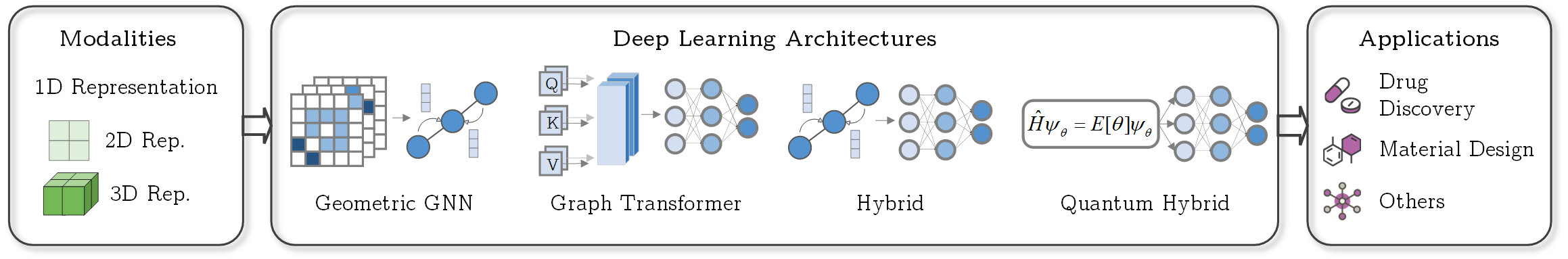}
    \caption{The pipeline of deep learning-driven MPP}
    \label{fig:pipeline}
\end{figure*}

\xsrv{The survey delineates a comprehensive pipeline for deep learning-driven MPP, as illustrated in Figure~\ref{fig:pipeline}. The pipeline comprises three principal stages: molecular data acquisition, deep learning model selection and training, and deployment for downstream analytical tasks.}

\xsrv{The pipeline initiates with molecular datasets of varying modalities, spanning from the simplest 1D representations to geometric conformations. This spectrum reflects distinct trade-offs between computational efficiency and spatial feature preservation. For instance, SMILES offers sequential processing advantages amenable to RNN and Transformer architectures, but SMILES may overlook stereochemical details, while molecular graphs capture topological relationships at higher computational cost.}

\xsrv{Following data acquisition, modality-specific preprocessing transforms molecular data into model-compatible formats. Specifically, 1D sequential representations undergo tokenization~\upcite{chemberta, smile2vec, MolBERT, daTransformer, spvec}, molecular graphs are featurized using atomic and bond descriptors~\upcite{dimenet, MPNN, Graphormer, mol2vec, mole, graphmae}, and 3D structures are voxelized for convolutional processing~\upcite{SchNet}. The preprocessed data are then embedded and fed into respective specialized architectures aligned with each modality: RNNs~\upcite{smile2vec} and Transformers~\upcite{chemberta, Graphormer, mole, MolBERT, daTransformer} for sequential data, GNNs~\upcite{dimenet, MPNN, graphmae} for graph structures, and CNNs~\upcite{SchNet} for grid-based inputs.}

\xsrv{The trained models are ultimately deployed to address MPP tasks. Key applications include the precise quantification of physicochemical properties and the prediction of ADMET endpoints~\upcite{smile2vec, mol2vec, wang2022molecular, MolBERT, graphmae, moleculenet}, encompassing absorption, distribution, metabolism, toxicity, and other safety profiles. The pipeline also facilitates applications in drug discovery, such as virtual screening for novel antibiotics, and in quantum chemistry for calculating molecular energies. This pipeline establishes deep learning as a versatile framework for advancing molecular property prediction.}

\subsection{Taxonomy}
\label{sec:taxonomy}
\xs{Formats of data consumed in supervised learning processes, deep learning models processing specific types of data, and analytical applications that specialized models target are vital factors determining analysis performances in MPP tasks. Therefore, this survey \rv{holistically} categorizes existing MPP approaches along three parallel dimensions as demonstrated in Figure \ref{fig:taxonomy}: modalities of input data, inherent architecture of the models employed, and molecular analytical applications.}

Organic molecules display vast structural diversity shaped by carbon connectivity, functional groups, and spatial conformation, all governing physicochemical behavior. To capture these features, molecular representations span multiple dimensions. One-dimensional encodings like SMILES enable scalable self-supervised learning but lack spatial awareness for geometry-sensitive tasks. Two- and three-dimensional graph and geometric formats better capture topology and conformation, improving quantum and stereochemical modeling at higher computational cost. This dimensional progression parallels evolving architectures. Sequential models handle symbolic dependencies, Transformers learn global context via self-attention, and GNNs achieve high structural sensitivity with linear scalability. Geometric networks incorporate 3D spatial features for quantum accuracy, while multimodal hybrids fuse complementary views for broader robustness. Spanning drug discovery, materials design, and general property prediction, these approaches collectively advance the goal of balancing representational fidelity, generalization, and computational efficiency in molecular AI.

\subsection{Operational Decision Framework}
\label{sec:decision_framework}
\rv{The taxonomy above organizes MPP methodologies descriptively. However, practitioners must select methods under concrete constraints including data availability, compute budget, 3D coordinate accessibility, out-of-distribution risk, and interpretability requirements. Table~\ref{tab:decision_matrix} distills the evidence presented throughout this survey into a constraint-driven decision matrix mapping common practical scenarios to recommended paradigm--architecture combinations.}

\rv{Several cross-cutting observations inform these recommendations. Descriptor-based models retain clear advantages in low-data and latency-critical settings, offering millisecond-level inference~\upcite{drug_ml} and interpretable features at the cost of limited extrapolation to novel chemotypes. When 3D geometry is available and properties are spatially sensitive (e.g., quantum energies, binding affinities), geometric deep learning methods achieve the highest accuracy. Within this paradigm, distance-only models (e.g., SchNet~\upcite{SchNet}) suffice for radially dominated properties, while directional models (e.g., DimeNet~\upcite{dimenet}, SphereNet~\upcite{spherenet}) are preferred for torsion-sensitive tasks. Foundation models offer the greatest gains when task-specific labels are scarce but large unlabeled corpora are available for pretraining~\upcite{zhou2023unimol, chemberta}, though at higher computational and interpretability cost.}

\rv{Crucially, our comparative analysis (Section~\ref{sec:comparative_analysis}) shows that method rankings are unstable across evaluation protocols: Four different models claim the best score across six ADME endpoints under temporal splitting (Table~\ref{tab:splitting}), in contrast to the compressed rankings observed on MoleculeNet (Table~\ref{tab:prediction_performance}). This argues against any single paradigm being universally optimal and supports task-adaptive selection validated under deployment-realistic evaluation protocols~\upcite{Ash2025Protocols}.}

\begin{table}[t]
\centering
\caption{Operational decision matrix for molecular property prediction}
\vspace{-0.15in}
\label{tab:decision_matrix}
\footnotesize
\setlength{\tabcolsep}{4pt}
\renewcommand{\arraystretch}{1.18}

{\renewcommand{\tabularxcolumn}[1]{m{#1}}%
\begin{tabularx}{\linewidth}{@{}
    >{\raggedright\arraybackslash}m{2.7cm}
    >{\raggedright\arraybackslash}m{2.2cm}
    >{\raggedright\arraybackslash}X
    >{\raggedright\arraybackslash}X
@{}}
\toprule
\textbf{Primary Constraint} &
\textbf{Recommended Paradigm} &
\textbf{Representative Methods} &
\textbf{Key Trade-offs} \\
\midrule

\makecell[l]{Low Compute\\ / Low Data} &
Descriptor ML &
RF/XGBoost + ECFP; N-Gram~\upcite{liu2019n} &
Fast (\(\sim\)ms/mol) and interpretable, but limited extrapolation to novel chemotypes. \\

\addlinespace[2pt]
\makecell[l]{3D-Sensitive Properties\\(Quantum, Docking)} &
Geometric GNNs &
DimeNet++~\upcite{gasteiger2020fast}; SphereNet~\upcite{spherenet}; EGNN~\upcite{EGNN} &
High accuracy on spatial properties, but requires conformer generation. \\

\addlinespace[2pt]
\makecell[l]{Scarce Labels\\ / High OOD Risk} &
\makecell[l]{Foundation Models\\ / Pretrained GNNs} &
\makecell[l]{MolE~\upcite{mole}; Uni-Mol~\upcite{zhou2023unimol}; ChemBERTa~\upcite{chemberta}; \\ ChemRL-GEM~\upcite{fang2022geometry}} &
Strong few-shot transfer, but high pretraining cost and reduced interpretability.\\

\addlinespace[2pt]
\makecell[l]{First-Principles Rigor\\(\(\sim\)0.1 kcal/mol)} &
\makecell[l]{Quantum\\ / Quantum Hybrid} &
\makecell[l]{DFT; Neural wavefunctions~\upcite{gao2023generalizing}; \\ PennyLane~\upcite{arrazola2021differentiable}} &
Near-chemical accuracy, but steep scaling limits system size. \\

\bottomrule
\end{tabularx}}
\vspace{-0.1in}
\end{table}
\section{Representation Modalities}
\label{sec:representation}

A molecule's representation fundamentally determines how a learning algorithm perceives underlying chemical information. As introduced in Section~\ref{sec:preliminaries}, the MPP task maps a molecular input to a target property; the choice of input modality directly governs the inductive bias, expressiveness, and data requirements of the downstream model. We categorize common representations by modality and dimensionality: \textbf{1D sequence-based encodings}, \textbf{2D topological graphs}, \textbf{3D geometric conformations}, \textbf{images and expert-crafted features}, and \textbf{multimodal combinations}. Each modality carries distinct advantages and limitations in MPP tasks.

\subsection{1D Representations}
\label{sec:1d-rep}
One-dimensional representations describe a molecule as a linear sequence of symbols. The most prevalent format is the SMILES string, with newer alternatives like SELFIES addressing some of SMILES' validity shortcomings. These encodings allow molecular structures to be treated as sentences, enabling the direct application of natural language processing (NLP) models to chemistry~\upcite{Alhmoudi2025}. Formally, a molecule $m$ is represented as a sequence $s = (c_1, c_2, \dots, c_L)$, where each $c_i$ comes from a vocabulary $\mathcal{V}$ of atom symbols, bond types, and structural tokens (e.g., C, N, $=$, branches). This format enables the direct use of sequence-based neural architectures~\upcite{smilesbert,MolBERT,chemberta}, including recurrent neural networks (RNNs) and Transformers. Sequence-based representations are attractive because they require no precomputed structure; large chemical databases often provide molecules only in SMILES or similar formats. Modern chemical language models leverage these 1D encodings to learn latent representations without the need for expensive 3D coordinates or expert features.

\begin{figure*}
    \centering
    \includegraphics[width=\linewidth]{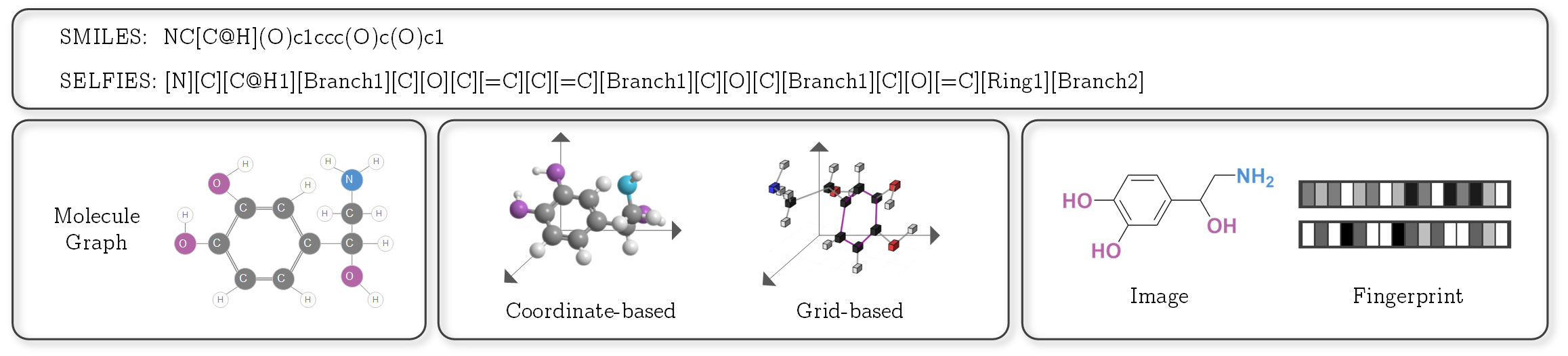}
    \caption{\rv{Overview of molecular representation modalities: 1D sequences, 2D graphs, 3D geometric conformations, and multimodal combinations.}}
    \label{fig:representation}
\end{figure*}

\textbf{SMILES}
\label{sec:smiles}
The Simplified Molecular Input Line Entry System (SMILES) encodes a chemical graph as a sequence of ASCII characters. SMILES strings are human-readable and compact, and they encode atoms, connectivity, and stereochemistry in a linear form. For example, ``CC(=O)O'' represents acetic acid.  \rv{A key benefit of SMILES is that enormous public databases (e.g., ZINC, PubChem) are available,} supporting large-scale training of ``chemical language models''~\upcite{Alhmoudi2025}. Transformers such as ChemBERTa~\upcite{chemberta} have been pretrained on tens of millions of unlabeled SMILES, achieving strong downstream performance on property prediction tasks.

However, SMILES has notable limitations. It is not canonical by default: a given molecule can be written in many different SMILES, depending on traversal order. Furthermore, arbitrary strings \rv{do not always correspond to valid molecules, which introduces invalid outputs during generation.} These issues can be mitigated by using canonical SMILES for consistency and augmenting models with randomized SMILES during training to improve robustness. Despite these limitations, SMILES remains widely used, and large self-supervised models have achieved performance rivaling graph-based methods.

\textbf{SELFIES}
\label{sec:selfies}
The Self-Referencing Embedded String (SELFIES)~\upcite{selfies} representation resolves SMILES' validity problem: every SELFIES string corresponds to a valid molecule by design. This is achieved through a context-free grammar enforcing valency and bonding constraints. Although SELFIES sequences are typically longer than their SMILES counterparts, the guaranteed validity greatly stabilizes molecular generation tasks. Recent studies demonstrate that SELFIES-based transformers (e.g., SELFormer~\upcite{selformer}) achieve comparable or superior performance to ChemBERTa on benchmarks like ESOL and SIDER, while avoiding invalid outputs. Extensions such as Group SELFIES further compress representations by grouping functional substructures into unique group tokens \rv{while preserving robustness}~\upcite{groupselfies}.

\textbf{Other Linear Encodings}
\label{sec:sequence}
Besides SMILES and SELFIES, other line notations exist. The IUPAC International Chemical Identifier (InChI)~\upcite{inchi} provides a canonical, layered text representation unique to each molecule, \rv{though less convenient for ML models due to its length and complex formatting.} DeepSMILES~\upcite{deepsmiles} modifies SMILES syntax to remove ring indices and improve parsing stability. In general, SMILES and SELFIES dominate 1D molecular modeling due to their balance of simplicity, expressiveness, and compatibility with transformer architectures.

\subsection{2D Molecular Graphs}
\label{sec:2d-rep}
In a 2D modality, a molecule is represented as an attributed graph $G = (V, E)$ capturing atomic connectivity. Each node $v_i \in V$ corresponds to an atom, and each edge $(v_i, v_j) \in E$ represents a bond. Initial node feature vectors $\mathbf{x}_v$ encode atomic properties (element, degree, aromaticity, hybridization), while edge features $\mathbf{x}_e$ encode bond order and stereochemistry~\upcite{MPNN,SchNet}. The adjacency matrix $A_{ij}$ or neighbor list encodes topology. GNNs operate on these structures through iterative message passing: at each layer, node embeddings are updated by aggregating local information from neighbors, typically as
\begin{equation}
\label{eq:message-passing}
h_i^{(k+1)} = \phi \big(h_i^{(k)}, \bigoplus_{j \in \mathcal{N}(i)} \psi(h_i^{(k)}, h_j^{(k)}, e_{ij}) \big),
\end{equation}
where $\psi$ and $\phi$ are learnable functions and $\bigoplus$ is a permutation-invariant aggregator (e.g., sum or mean). After $K$ iterations, a global readout function (sum, mean, attention pooling) produces a molecule-level embedding for property prediction. This framework generalizes handcrafted fingerprints like ECFP by learning optimal feature propagation rules. Architectural variants and extensions of this message-passing framework are discussed in Section~\ref{sec:gnn}.

Graph representations align closely with chemical intuition and are invariant to atom indexing. Modern variants incorporate higher-order interactions (angles, triplets) and physics-inspired priors. However, 2D graphs ignore conformational variability and long-range spatial effects. Despite this, GNNs remain highly competitive, especially for properties determined primarily by bonding topology (e.g., solubility, toxicity).

\subsection{3D Geometric Representations}
\label{sec:3d-rep}
Three-dimensional representations describe molecules using atomic coordinates $\mathbf{r}_i \in \mathbb{R}^3$ alongside atomic numbers $Z_i$. They are important for modeling quantum and spatial effects. Properties such as binding affinity, chemical reactivity, and spectroscopic characteristics are fundamentally influenced by the spatial arrangement of atoms~\upcite{SchNet,EGNN,fuchs2020se,spherenet,tholke2022torchmd,zhou2023unimol}. Two major families exist: grid-based encodings and coordinate-based encodings. Grid-based encodings are particularly common in structure-based modeling settings (e.g., protein–ligand complexes), whereas coordinate-based encodings dominate small-molecule quantum and property prediction tasks.

\textbf{Grid-Based Encodings}
The molecule is embedded in a 3D voxel grid, where each voxel stores features such as atomic densities or electrostatic potentials. 3D convolutional neural networks (3D-CNNs) can then learn local spatial patterns. While conceptually straightforward, voxelization is often memory-intensive and inefficient due to the sparsity of molecular data in 3D space.

\textbf{Coordinate-Based Encodings}
Coordinate-based models take atomic positions and types $\{\mathbf{r}_i, Z_i\}$ directly as input. A fundamental physical requirement is that predictions of scalar molecular properties must be invariant to global translations, rotations, and reflections of the input coordinates, a constraint known as E(3) invariance. Models enforce this either by constructing features exclusively from geometric invariants, or by maintaining equivariant intermediate representations that transform predictably under spatial operations and are contracted to invariant scalars via a symmetric readout. Formal definitions and transformation laws are provided in Appendix~\ref{app:symmetry}.

Modern coordinate-based encodings capture molecular geometry at increasing levels of angular resolution. Distance-based features use pairwise distances $d_{ij} = \|\mathbf{r}_i - \mathbf{r}_j\|$ expanded via radial basis functions; directional encodings additionally incorporate bond angles $\theta_{ijk}$ and dihedral angles $\phi_{ijkl}$; and spherical-basis representations combine radial functions with spherical harmonics $Y_\ell^m$ to jointly encode distance and orientation~\upcite{spherenet}. Detailed formulations, including radial basis expansions and angle definitions, are given in Appendix~\ref{app:symmetry}.

Representative architectures illustrate the breadth of this design space. EGNN~\upcite{EGNN} maintains equivariant atomic embeddings by propagating geometric messages that transform consistently under E(3) operations. DimeNet~\upcite{dimenet} achieves E(3) invariance by embedding directional triplet interactions $(i,j,k)$ via spherical Bessel functions and spherical harmonics, directly encoding both radial and angular information into invariant features. SphereNet~\upcite{spherenet} extends this to torsion angles, incorporating dihedral interactions to capture finer geometric detail. Collectively, these architectures~\upcite{dimenet,spherenet} have reported near-chemical-accuracy performance on established benchmarks such as QM9 and MD17, while offering much lower inference cost than repeated quantum-chemical calculations once trained.

Challenges include the computational cost of conformer generation and the ambiguity introduced by multiple low-energy conformations per molecule. Despite these, 3D-aware models are often essential for tasks sensitive to molecular shape and electrostatics (e.g., docking, reaction energetics).

\subsection{Images and Expert-Crafted Features}
\label{sec:image-rep}
Beyond the primary learned representations discussed above, molecules can also be processed as 2D or 3D images by rendering their structures into pixel-based formats, enabling the use of convolutional neural networks (CNNs).

Expert-crafted features remain widely used as complementary inputs. Molecular descriptors are numerical vectors $\mathbf{d} \in \mathbb{R}^k$ quantifying molecular structure and behavior, including topological descriptors (connectivity patterns), electronic and geometric descriptors, and physicochemical descriptors. Molecular fingerprints~\upcite{scfp} are binary vectors $F \in \{0, 1\}^k$ encoding substructural patterns via key-based or hash-based schemes. These hand-crafted features are often combined with learned representations in multimodal deep learning models, as discussed below.

\subsection{Multimodal Representations}
\label{sec:mult-rep}
Multimodal approaches combine multiple molecular representations to capture complementary information. For example, a SMILES string encodes syntactic patterns, a graph captures connectivity, and 3D coordinates capture spatial structure. Fusion can occur at the feature, latent, or decision level.

Early works concatenated handcrafted descriptors with graph embeddings~\upcite{yang2019analyzing}. Recent architectures adopt learned fusion: GraSeq~\upcite{guo2020graseq} jointly trains SMILES and graph encoders, while MoL-MoE~\upcite{zhang2024mol} integrates SMILES, SELFIES, and graph experts via a mixture-of-experts framework. The gating network adaptively weighs modalities per task, achieving state-of-the-art results on MoleculeNet~\upcite{moleculenet}. Similar multimodal strategies have been extended to text and image modalities (e.g., spectra or microscopy), creating unified molecular foundation models.

While multimodal fusion improves robustness and generalization, it introduces complexity and requires all modalities to be available at inference. Careful regularization and interpretability analyses are needed to ensure that improvements stem from complementary information rather than redundancy.

\section{Model Architectures}
\label{sec:architectures}

This section examines the evolutionary trajectory of architectural paradigms for molecular property prediction, tracing their development from foundational concepts to contemporary innovations. We analyze how each approach emerged to address limitations in molecular representation, highlighting key breakthroughs that defined their development paths.
\begin{figure*}
    \centering
    \includegraphics[width=\linewidth]{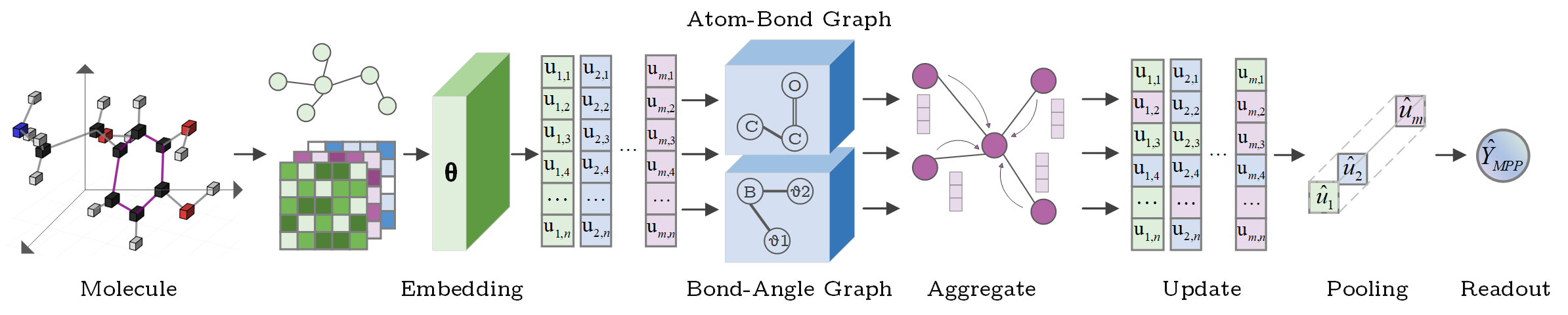}
    \caption{Geometric GNN in MPP}
    \label{fig:gnn}
\end{figure*}

\subsection{Geometric Graph Neural Networks}
\label{sec:gnn}
The evolution of geometric GNNs began with early attempts to incorporate spatial information into graph representations. Initial approaches like SchNet~\upcite{SchNet} pioneered distance-based modeling by encoding pairwise atomic distances through radial basis functions. These early models demonstrated that simple distance features could capture essential quantum properties but struggled with angular dependencies. \rv{This limitation spurred the development of directional architectures, with DimeNet~\upcite{dimenet} introducing spherical harmonics to model bond angles and dihedral angles. This breakthrough enhanced the accuracy of predicting torsion-sensitive properties and demonstrated strong predictive performance for other molecular properties. Gasteiger et al.~\upcite{dimenet} reported that DimeNet outperforms previous GNNs on average by 76\% on MD17 and by 31\% on QM9.}

The field progressed toward more sophisticated symmetry-aware models with the introduction of Tensor Field Networks~\upcite{thomas2018tensor} that first implemented spherical irreducible representations. This foundation enabled the SE(3)-Transformer~\upcite{fuchs2020se}, which established the modern paradigm of equivariant feature updates using Wigner-D matrices. These vector-based architectures could maintain strict rotation/translation invariance while updating atomic positions. \rv{More recent models} like GemNet~\upcite{gasteiger2021gemnet} incorporate explicit quadruple interactions and Allegro~\upcite{nomura2025allegro} implements Bessel function projections that better capture electron density distributions.

Current geometric GNNs~\upcite{SchNet,gasteiger2020fast,han2022directed,shen2023molecular} achieve near-density functional theory accuracy on quantum chemical properties including formation energies and molecular forces. Their development represents a continuous refinement of symmetry preservation mechanisms, from simple distance encoding to sophisticated equivariant operations. \rv{Nevertheless, such architectures face constraints stemming from their computational demands on large molecules and their requirement for precise geometric information, prompting the emergence of alternative methodological frameworks.}

\subsection{Graph Transformers}
\label{sec:transformer}
Graph Transformers~\upcite{shehzad2024graph} emerged as a response to the local receptive field limitations of early GNNs. The initial breakthrough came with the adaptation of NLP transformers to graph structures in Graph Attention Networks~\upcite{velickovic2018graph}, which implemented attention-based neighborhood aggregation but lacked global context. This limitation was first addressed in Graph-BERT~\upcite{jha2023graph}, which applied full self-attention to molecular graphs but ignored spatial relationships. The pivotal innovation arrived with Graphormer~\upcite{Graphormer}, which introduced structural encoding of shortest path distances and spatial relationships, \rv{enabling Transformers on graphs. On the large-scale quantum chemistry regression dataset OGB-LSC, Graphormer achieves over a 10 percentage-point reduction in relative error compared to most mainstream GNN variants~\upcite{Graphormer}, demonstrating its capability on large-scale graphs.}

The subsequent evolution focused on geometric integration, with Geometric Transformers~\upcite{fuchs2020se} incorporating 3D coordinates directly into attention calculations through continuous kernel methods. This period saw the rise of equivariant attention mechanisms in models like Equivariant Graph Attention Transformer~\upcite{liao2023equiformer}, which ensures rotational equivariance by operating on irrep features and using equivariant operations throughout the Transformer blocks. 
\rv{Modern graph transformer architectures that incorporate global self-attention have shown strong performance in predicting electronic properties involving delocalized electrons and $\pi$-stacking phenomena, often providing improved accuracy over standard message-passing GNN baselines~\upcite{Graphormer,rong2020self}.} Their evolution reflects an ongoing tension between modeling completeness and computational feasibility - a challenge that continues to drive architectural innovation as researchers seek to balance global context with practical scalability.

\begin{figure*}
    \centering
    \includegraphics[width=\linewidth]{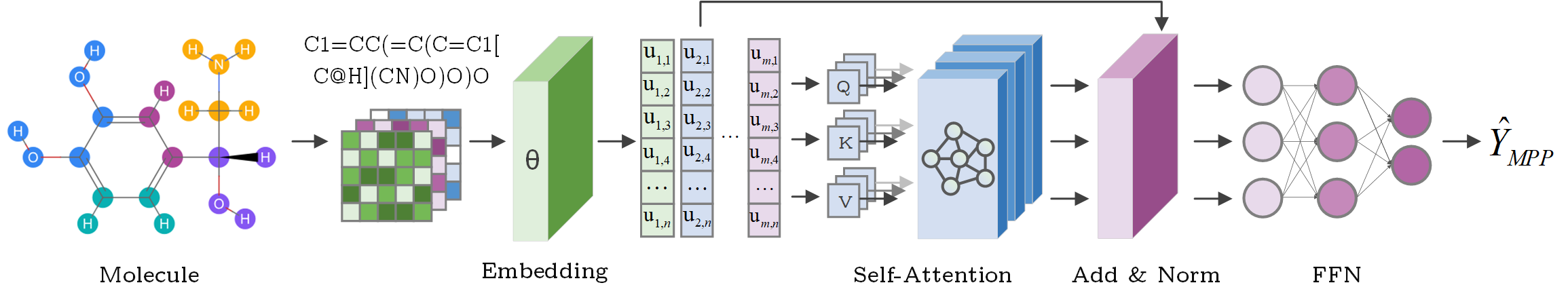}
    \caption{Transformer in MPP}
    \label{fig:transformer}
\end{figure*}

\subsection{Hybrid Architectures}
\label{sec:hybrid}
Hybrid architectures originated from efforts to reconcile the complementary strengths of message-passing networks and attention mechanisms. Early attempts simply concatenated GNN and transformer outputs, yielding marginal improvements. A key step forward was the emergence of hybrid MPNN/Transformer models such as GPS++~\upcite{masters2022gps++}, which run expressive message passing and structurally (or 3D-) biased self-attention in parallel, then fuse the two streams via a simple summation followed by an MLP, while leveraging 3D geometric cues and an auxiliary denoising objective. This supports a common three-component recipe: local feature extraction via MPNNs, global context modeling via (biased) attention, and learned fusion of the two representations.

The subsequent evolution focused on geometric integration, with models like TorchMD-NET~\upcite{tholke2022torchmd} incorporating physical gradients directly into attention mechanisms, enabling simultaneous prediction of energies and atomic forces. \rv{This phase introduced dynamic gating mechanisms} in architectures such as Moleformer~\upcite{yuan2023molecular}, which employs attention-based routing \rv{between local and global processing pathways}. The current innovation wave centers on efficiency, with GraphGPS~\upcite{rampavsek2022recipe} pioneering subgraph sampling for linear complexity and MambaMixer~\upcite{behrouz2024mambamixer} integrating state-space models for long-sequence molecular modeling.

Contemporary hybrids represent the most balanced approach, achieving state-of-the-art results on comprehensive benchmarks like PCQM4Mv2~\upcite{lu2023highly}. \rv{Moreover, on PCQM4Mv2, GPS++~\upcite{masters2022gps++} reports single-model accuracy competitive with the strongest prior Transformer baselines while using substantially fewer parameters, and its final ensemble achieved first place in the OGB-LSC 2022 challenge.} Their development reflects increasing sophistication in feature fusion strategies and computational optimization - positioning them as leading candidates for molecular foundation models. The ongoing challenge remains developing scalable unified architectures that seamlessly integrate physical priors without compromising computational efficiency.

\subsection{Quantum Hybrid Approaches}
\label{sec:quantum}
Quantum hybrid architectures emerged from early attempts to incorporate physical constraints into neural networks. Initial physics-informed neural networks~\upcite{razakh2021pnd} applied simple quantum mechanical regularizers but showed limited accuracy gains. The breakthrough came with Hamiltonian-informed architectures~\upcite{gerard2022gold} that implemented strict Schrödinger equation constraints through specialized loss functions, reducing errors on challenging transition metal complexes. This established the paradigm of direct physics enforcement.

The field progressed toward more integrated approaches with neural wavefunctions~\upcite{gao2023generalizing}, which parameterized quantum states through constrained neural networks with Jastrow factors. Concurrently, density functional networks~\upcite{dick2020neuralxc} began replacing exchange-correlation functionals with learnable neural representations. The most significant advancement arrived with differentiable quantum solvers~\upcite{arrazola2021differentiable} that implemented end-to-end trainable Kohn-Sham equations using neural operators.

Current innovations focus on computational feasibility, with Fourier Neural Operators~\upcite{li2020fourier} solving quantum equations in spectral space and Orbital Networks predicting molecular orbitals through symmetry-adapted representations. These approaches achieve unprecedented accuracy ($<$0.1 kcal/mol) on challenging quantum properties but require specialized optimization techniques to maintain self-consistent convergence. Their evolution represents the frontier of physics-grounded molecular modeling, though computational demands continue to limit application to small molecules. The ongoing challenge is developing scalable quantum-classical frameworks for foundation model pretraining.

\section{Datasets and Benchmarks}
\label{sec:benchmarks}

\subsection{Benchmark Datasets}
\label{sec:benchmark_datasets}
\rv{To advance research in molecular property prediction, benchmark datasets play a central role in the development, evaluation, and comparison of deep learning models. Among these, MoleculeNet~\upcite{moleculenet} has become one of the most widely adopted benchmark suites, curating 16 datasets across four categories: quantum mechanics (QM7--QM9), physical chemistry (ESOL, FreeSolv, Lipophilicity), biophysics (BACE, HIV, MUV, PCBA, PDBbind), and physiology (BBBP, Tox21, ToxCast, ClinTox, SIDER). These datasets collectively span regression and classification tasks over quantum-mechanical, physicochemical,
pharmacological, and toxicological domains, and have served as a common reference point for comparing molecular representations and learning algorithms.
Table~\ref{tab:benchmark_dataset} summarizes the specifications for each dataset, including target properties, task types, dataset sizes, recommended splits, and evaluation metrics.}

\rv{However, despite the widespread adoption of MoleculeNet~\upcite{moleculenet} as a \textit{de facto} standard for benchmarking molecular property prediction models, several critical shortcomings have been identified that undermine the validity of conclusions drawn from such comparisons~\upcite{walters2023benchmarks}. First, many constituent datasets suffer from data curation issues, including invalid SMILES representations, inconsistent chemical standardization, and pervasive undefined stereochemistry. For example, 71\% of molecules in the BACE dataset contain at least one undefined stereocenter, making it unclear what chemical entity is actually being modeled. Second, datasets aggregated from multiple literature sources (e.g., the BACE set compiled from 55 separate papers) lack consistent experimental protocols, and combining IC$_{50}$ values measured under heterogeneous assay conditions introduces substantial noise that may exceed the signal one hopes to model. Third, several benchmark endpoints are poorly suited for meaningful method comparison: the ESOL solubility dataset spans over 13 log units, far exceeding the 2--3 log dynamic range encountered in pharmaceutical practice, thereby inflating apparent model performance, while complex physiological endpoints such as blood--brain barrier penetration and clinical toxicity involve multifactorial mechanisms that are difficult to capture with binary labels derived from heterogeneous sources. Fourth, datasets like the HIV set contain a high prevalence of compounds likely to generate assay artifacts, with 70\% of confirmed actives triggering structural alerts. These issues collectively argue that the field urgently needs higher-quality, carefully curated benchmarks. These must rely on simple, reproducible endpoints, including controlled measurements of aqueous solubility, membrane permeability, and metabolic stability. Furthermore, explicitly defined chemical structures and standard train/test splits should be used~\upcite{walters2023benchmarks, Fang2023Prospective}. Efforts such as the publicly released ADME dataset from Fang et al.~\upcite{Fang2023Prospective}, which provides consistently measured data on commercially available drug-like compounds across six \textit{in vitro} ADME endpoints, represent a promising step in this direction.}

\subsection{Dataset Splitting Strategies}
\label{sec:dataset_split}
\rv{The choice of splitting strategy is a critical yet often underappreciated factor in model evaluation. Random splitting frequently results in highly similar molecules appearing in both training and test sets, leading to overly optimistic performance estimates~\upcite{sheridan2004similarity}. Scaffold splitting mitigates this to some extent, but has a well-documented limitation: structurally similar molecules may possess distinct scaffolds, allowing near-trivial predictions to leak into the test set~\upcite{guo2024scaffold}. The systematic evaluation by Guo et al.~\upcite{guo2024scaffold,guo2025umap} across 60 NCI-60 cancer cell line datasets demonstrated that scaffold splits significantly overestimate virtual screening performance compared to UMAP-based splits, and that model selection based on scaffold-split results can lead to suboptimal prospective outcomes. However, Sheridan~\upcite{sheridan2013timesplit} showed early on that splitting strategies yielding greater train--test dissimilarity consistently produce lower, overly pessimistic performance estimates, whereas time-split cross-validation produces the most realistic estimates of prospective prediction. Regarding UMAP-based methods specifically, Walters~\upcite{walters2024splitting} raised two concerns. First, the UMAP-based splits rely on a projection that captures only a portion of the structural similarity present in the dataset. Second, agglomerative clustering often produces highly imbalanced cluster sizes, which may artificially increase the variability of the evaluation metrics. In practice, time-based splits are considered the gold standard, but most public benchmarks lack the necessary timestamps. Overall, the community is moving toward more stringent splitting protocols that better reflect the genuine difficulty of predicting properties for structurally novel compounds encountered in real-world drug discovery campaigns.}
\subsection{Protocol Selection}
\label{sec:protocol_selection}
\rv{Beyond benchmark datasets and splitting strategies, the evaluation protocol itself largely determines whether reported improvements are credible. Ash et al.~\upcite{Ash2025Protocols} argue that molecular property prediction studies often suffer from a replicability gap: comparisons typically report a single mean score with no hypothesis testing, yet drug discovery datasets are frequently small ($\leq 10^4$), imbalanced, and noisy, making observed performance differences highly sensitive to which compounds fall into training versus test partitions.}

\rv{To capture this variability, Ash et al.~\upcite{Ash2025Protocols} recommend $5\times5$ repeated cross-validation as a practical default for datasets of 500--100{,}000 molecules, producing 25 per-fold performance samples suitable for statistical testing. Although repeated resampling introduces some inter-iteration dependence, this overlap is less problematic than that of vanilla 10-fold CV, repeated random sampling, or bootstrapping, all of which exhibit elevated false positive rates or strong inter-sample dependence in simulation~\upcite{dietterich1998approximate, bates2024cv}. The same splits should be applied to all methods, enabling paired testing. The framework accommodates group-structured splits (scaffold- or similarity-cluster-based) via grouped CV utilities (e.g., \texttt{GroupKFold}), provided groups are non-overlapping and roughly balanced. Temporal splits, however, require chronology-preserving protocols, as discretizing time into arbitrary groups undermines the prospective realism that motivates temporal evaluation~\upcite{sheridan2013timesplit}. When hyperparameter tuning is needed, splitting each outer-training fold into training/validation subsets avoids the computational cost and metric constraints of full nested CV~\upcite{bates2024cv, Ash2025Protocols}.}

\rv{For statistical comparison, Ash et al.~\upcite{Ash2025Protocols} recommend repeated-measures ANOVA followed by Tukey's HSD test, which accounts for the paired structure and corrects for multiple comparisons in a single procedure; for two-method comparisons this reduces to a paired $t$-test. When multiple metrics are assessed, a Bonferroni correction at the ANOVA stage maintains overall type~I error control.}

\rv{Statistical significance, however, does not ensure practical utility. As discussed in Section~\ref{sec:benchmark_datasets}, the ESOL dataset's 13-log dynamic range inflates apparent performance ($R^2$ drops from 0.68 to 0.33 over a realistic 3-log subrange)~\upcite{Ash2025Protocols, walters2023benchmarks}. Decisional impact metrics such as recall@precision or TNR@recall better reflect compound prioritization decisions~\upcite{Ash2025Protocols}. It is also informative to contextualize results with a lower bound from null models and an upper bound estimated from experimental variability. For method comparison studies, Ash et al. further recommend simultaneous confidence interval plots or MCSim plots as alternatives to single-value leaderboards for conveying uncertainty and effect size~\upcite{Ash2025Protocols}.}
\begin{table*}[!ht]
    \centering
    \caption{Summary of benchmark datasets for molecular \rv{property prediction}}
    \vspace{-0.15in}
    \label{tab:benchmark_dataset}
    \vspace{0.1cm}
    \renewcommand{\arraystretch}{1.2}
    \setlength{\tabcolsep}{8pt}
    \resizebox{\textwidth}{!}{
    \begin{tabular}{lcccccccc}
        \toprule[1.2pt]
        \textbf{Categories} & \textbf{Property} & \textbf{Dataset} & \textbf{Data type} & \textbf{Tasks} & \textbf{Type} & \textbf{Compounds} & \textbf{Rec-Splitting} & \textbf{Rec-Metric}\\
        \midrule[0.8pt]
        \multirow{4}{*}{Quantum mechanics}
        & Quantum mechanics  & QM7 & SMILES \& 3D coordinates & 1 & Regression & 7,165 & Stratified & MAE \\
        & Quantum mechanics  & QM7b & 3D coordinates & 14 & Regression & 7,211 & Random & MAE \\
        & Quantum mechanics  & QM8 & SMILES \& 3D coordinates & 12 & Regression & 21,786 & Random & MAE \\
        & Quantum mechanics  & QM9 & SMILES \& 3D coordinates & 12 & Regression & 133,885 & Random & MAE \\
        \midrule
        \multirow{3}{*}{Physical chemistry}
        & Solubility  & ESOL & SMILES & 1 & Regression & 1,128 & Random & RMSE  \\
        & Solv. Energy & \rv{FreeSolv} & SMILES & 1 & Regression & 643 & Random & RMSE \\
        & Hydrophobicity, Hydrophilicity & Lipophilicity & SMILES & 1 & Regression & 4,200 & Random & RMSE  \\
        
        \midrule
        \multirow{5}{*}{Physiology}
        & \rv{Permeability} & BBBP & SMILES & 1 & Classification & 2,053 & Scaffold & ROC-AUC \\
        & Toxicity & Tox21 & SMILES & 12 & Classification & 8,014 & Random & ROC-AUC \\
        & Toxicity & ToxCast & SMILES & 617 & Classification & 8,615 & Random & ROC-AUC \\
        & Toxicity & ClinTox & SMILES & 2 & Classification & 1,491 & Random & ROC-AUC \\
        & Side-effects & SIDER & SMILES & 27 & Classification & 1,427 & Random & ROC-AUC \\
        \midrule
        \multirow{5}{*}{Biophysics}
        & Biological Affinity & BACE & SMILES & 1 & Classification & 1,522 & Scaffold & ROC-AUC \\
        & Efficacy & HIV & SMILES & 1 & Classification & 41,913 & \rv{Scaffold} & ROC-AUC \\
        & Bioactivity & PCBA & SMILES & 128 & Classification & 439,863 & Random & PRC-AUC \\
        & Bioactivity & MUV & SMILES & 17 & Classification & 93,127 & Random & PRC-AUC \\
        & Binding affinity & PDBbind & SMILES \& 3D coordinates & 1 & Regression & 11,908 & Time & RMSE \\
        
        \midrule
        \rv{\multirow{6}{*}{ADME}}
        & \rv{HLM (Human Liver Microsomal)} & \rv{ADME} & \rv{SMILES} & \rv{1} & \rv{Regression} & \rv{3,087} & \rv{Time} & \rv{Pearson R} \\
        & \rv{RLM (Rat Liver Microsomal)} & \rv{ADME} & \rv{SMILES} & \rv{1} & \rv{Regression} & \rv{3,054} & \rv{Time} & \rv{Pearson R} \\
        & \rv{MDR1-MDCK ER (Efflux Ratio)} & \rv{ADME} & \rv{SMILES} & \rv{1} & \rv{Regression} & \rv{2,642} & \rv{Time} & \rv{Pearson R} \\
        & \rv{Solubility} & \rv{ADME} & \rv{SMILES} & \rv{1} & \rv{Regression} & \rv{2,173} & \rv{Time} & \rv{Pearson R} \\
        & \rv{hPPB (Human Plasma Protein Binding)} & \rv{ADME} & \rv{SMILES} & \rv{1} & \rv{Regression} & \rv{1,808} & \rv{Time} & \rv{Pearson R} \\
        & \rv{rPPB (Rat Plasma Protein Binding)} & \rv{ADME} & \rv{SMILES} & \rv{1} & \rv{Regression} & \rv{885} & \rv{Time} & \rv{Pearson R} \\
        
        \bottomrule[1.2pt]
    \end{tabular}}
    
    \vspace{-0.1in}

    \begin{flushleft}
    {\scriptsize \rv{$^{\dagger}$ ADME dataset: HLM = Human Liver Microsomal Stability; RLM = Rat Liver Microsomal Stability; ER = MDR1-MDCK Efflux Ratio; Solubility = Solubility at pH 6.8; hPPB = Human Plasma Protein Binding; rPPB = Rat Plasma Protein Binding. Sample counts reflect valid (non-missing) values per endpoint.}}
    \end{flushleft}
\end{table*}
\subsection{Comparative Analysis of Approaches}
\label{sec:comparative_analysis}
\rv{Tables~\ref{tab:prediction_performance} and~\ref{tab:splitting} reveal instructive contrasts between model performance under scaffold splitting on MoleculeNet~\upcite{moleculenet} and time-based splitting on the ADME benchmark~\upcite{Fang2023Prospective}. Results in Table~\ref{tab:prediction_performance} are taken from KA-GAT\upcite{li2025kolmogorov}, and those in Table~\ref{tab:splitting} are from our own experiments (see Appendix~\ref{app:hyperparams} for hyperparameter settings).}

\rv{On MoleculeNet (Table~\ref{tab:prediction_performance}), the KA-GCN and KA-GAT variants~\upcite{li2025kolmogorov} achieve the strongest overall results, claiming the top position on six of seven datasets. Methods incorporating richer representations generally outperform simpler message-passing baselines such as D-MPNN~\upcite{yang2019analyzing} and AttentiveFP~\upcite{xiong2020pushing}, whether through geometry-enhanced pretraining (GEM~\upcite{fang2022geometry}, Uni-Mol~\upcite{zhou2023unimol}), contrastive learning (MolCLR~\upcite{wang2022molecular}, GraphMVP~\upcite{liu2021pre}), or expressive activation functions (KA-GCN, KA-GAT~\upcite{li2025kolmogorov}). Performance differences across methods are often modest, with many approaches clustering within a few percentage points of ROC-AUC.}

\rv{Under time-based splitting on the ADME benchmark~\upcite{Fang2023Prospective} (Table~\ref{tab:splitting}), however, two notable shifts emerge. First, absolute performance drops substantially, with Pearson~$R$ values rarely exceeding 0.7, reflecting the genuine difficulty of temporally out-of-distribution prediction. Second, relative rankings change markedly: AttentiveFP~\upcite{xiong2020pushing}, which ranks near the bottom on most MoleculeNet tasks, leads on RLM and ranks second on HLM. No single method dominates overall. Four different models claim the best score across the six ADME endpoints.}

\rv{This discrepancy carries practical implications. The compressed, relatively stable rankings observed on MoleculeNet under scaffold splitting may partly reflect the data leakage and metric inflation discussed in Sections~\ref{sec:benchmark_datasets} and~\ref{sec:dataset_split}. Using time-based splitting and more practical datasets, which better approximate prospective drug discovery, performance gaps widen and rankings become less stable, consistent with prior analyses~\upcite{sheridan2013timesplit, guo2024scaffold}. Without the rigorous statistical testing advocated by Ash et al.~\upcite{Ash2025Protocols}, it remains unclear whether many observed pairwise differences reflect genuine methodological improvements. These findings argue for task-adaptive modeling strategies in applied settings and reinforce the need to align benchmarking practices more closely with real-world evaluation conditions.}

\begin{table*}[tb]
	\centering
	\scriptsize
	\caption{Molecular property prediction performance. All baseline results on MoleculeNet~\upcite{moleculenet} datasets (BBBP, Tox21, ClinTox, SIDER, BACE, HIV, MUV) are \srv{reported in Li et al.}~\upcite{li2025kolmogorov} and evaluated using the ROC-AUC metric. A dash (–) indicates that the result was not reported in the original source or that the metric is not applicable. Standard deviations computed across multiple independent training runs are shown as subscripts. \textbf{Bold} fonts denote the best results and \underline{underlined} ones denote the second best.}
	\label{tab:prediction_performance}
    \vspace{-0.15in}
    \newcommand{\mstd}[2]{#1\textsubscript{$\pm$#2}}
	\begin{tabular}{@{}>{\raggedright\arraybackslash}p{2.5cm}*{7}{>{\centering\arraybackslash}p{1.1cm}}@{}}
		\toprule
		\rowcolor{white!10}
		\textbf{Method} & \textbf{BBBP$\uparrow$} & \textbf{Tox21$\uparrow$} & \textbf{ClinTox$\uparrow$} & \textbf{SIDER$\uparrow$} & \textbf{BACE$\uparrow$} & \textbf{HIV$\uparrow$} & \textbf{MUV$\uparrow$} \\
		\midrule

		D-MPNN~\upcite{yang2019analyzing}     &
		\mstd{0.710}{0.003} & \mstd{0.759}{0.007} & \mstd{0.906}{0.007} & \mstd{0.570}{0.007} & \mstd{0.809}{0.006} & \mstd{0.771}{0.005} & \mstd{0.786}{0.014} \\

		AttentiveFP~\upcite{xiong2020pushing} &
		\mstd{0.663}{0.018} & \mstd{0.781}{0.005} & \mstd{0.847}{0.003} & \mstd{0.606}{0.032} & \mstd{0.784}{0.022} & \mstd{0.757}{0.014} & \mstd{0.786}{0.015} \\

		N-GramRF~\upcite{liu2019n}            &
		\mstd{0.697}{0.006} & \mstd{0.743}{0.009} & \mstd{0.775}{0.040} & \mstd{0.668}{0.007} & \mstd{0.779}{0.015} & \mstd{0.772}{0.004} & \mstd{0.769}{0.002} \\

		N-GramXGB~\upcite{liu2019n}           &
		\mstd{0.691}{0.008} & \mstd{0.758}{0.009} & \mstd{0.875}{0.027} & \mstd{0.655}{0.007} & \mstd{0.791}{0.013} & \mstd{0.787}{0.004} & \mstd{0.748}{0.002} \\

		PretrainGNN~\upcite{hu2020strategies} &
		\mstd{0.687}{0.013} & \mstd{0.781}{0.006} & \mstd{0.726}{0.015} & \mstd{0.627}{0.008} & \mstd{0.845}{0.007} & \mstd{0.799}{0.007} & \mstd{0.813}{0.021} \\

		GraphMVP~\upcite{liu2021pre}          &
		\mstd{0.724}{0.016} & \mstd{0.759}{0.005} & \mstd{0.791}{0.028} & \mstd{0.639}{0.012} & \mstd{0.812}{0.009} & \mstd{0.770}{0.012} & \mstd{0.777}{0.006} \\

		MolCLR~\upcite{wang2022molecular}     &
		\mstd{0.722}{0.021} & \mstd{0.750}{0.002} & \mstd{0.912}{0.035} & \mstd{0.589}{0.014} & \mstd{0.824}{0.009} & \mstd{0.781}{0.005} & \mstd{0.796}{0.019} \\

		GEM~\upcite{fang2022geometry}         &
		\mstd{0.724}{0.004} & \mstd{0.781}{0.001} & \mstd{0.901}{0.013} & \mstd{0.672}{0.004} & \mstd{0.856}{0.011} & \mstd{0.806}{0.009} & \mstd{0.817}{0.005} \\

		Mol-GDL~\upcite{shen2023molecular}    &
		\mstd{0.728}{0.019} & \mstd{0.794}{0.005} & \mstd{0.966}{0.002} & \mstd{0.831}{0.002} & \mstd{0.863}{0.019} & \mstd{0.808}{0.007} & \mstd{0.675}{0.014} \\

		Uni-mol~\upcite{zhou2023unimol}       &
		\mstd{0.729}{0.006} & \mstd{0.796}{0.005} & \mstd{0.919}{0.018} & \mstd{0.659}{0.013} & \mstd{0.857}{0.002} & \mstd{0.808}{0.003} & \mstd{0.821}{0.013} \\

		SMPT~\upcite{li2024pre}               &
		\mstd{0.734}{0.003} & \mstd{0.797}{0.001} & \mstd{0.927}{0.002} & \mstd{0.676}{0.050} & \mstd{0.873}{0.015} & \mstd{0.812}{0.001} & \underline{\mstd{0.822}{0.008}} \\

		GNN-SKAN~\upcite{li2025gnn}           &
		\mstd{0.676}{0.014} & \mstd{0.747}{0.005} & --                  & \mstd{0.614}{0.005} & \mstd{0.747}{0.009} & \mstd{0.786}{0.015} & --                  \\

		GraphKAN~\upcite{zhang2024graphkan}   &
		\mstd{0.731}{0.017} & \mstd{0.753}{0.007} & \mstd{0.984}{0.003} & \mstd{0.837}{0.001} & \mstd{0.823}{0.011} & \mstd{0.711}{0.016} & \mstd{0.715}{0.014} \\

		KA-GNNs~\upcite{bresson2024kagnns}    &
		\mstd{0.721}{0.003} & \mstd{0.730}{0.012} & \mstd{0.972}{0.001} & \mstd{0.831}{0.004} & \mstd{0.752}{0.011} & \mstd{0.717}{0.018} & \mstd{0.701}{0.006} \\

		KA-GCN~\upcite{li2025kolmogorov}      &
		\textbf{\mstd{0.787}{0.014}} & \underline{\mstd{0.799}{0.005}} & \textbf{\mstd{0.992}{0.005}} & \underline{\mstd{0.842}{0.001}} &
		\textbf{\mstd{0.890}{0.014}} & \underline{\mstd{0.821}{0.005}} & \textbf{\mstd{0.834}{0.009}} \\

		KA-GAT~\upcite{li2025kolmogorov}      &
		\underline{\mstd{0.785}{0.021}} & \textbf{\mstd{0.800}{0.006}} & \underline{\mstd{0.991}{0.005}} & \textbf{\mstd{0.847}{0.002}} &
		\underline{\mstd{0.884}{0.004}} & \textbf{\mstd{0.823}{0.002}} & \textbf{\mstd{0.834}{0.010}} \\

		\bottomrule
	\end{tabular}
	\vspace{-0.05in}
\end{table*}

\begin{table*}[tb]
	\centering
	\footnotesize
	\caption{\rv{Performance comparison of different models on six ADME~\upcite{Fang2023Prospective} datasets with \textbf{time-based} splitting. \srv{All results are from our own experiments using a unified molecular encoding scheme (see Appendix~\ref{app:hyperparams} for hyperparameter settings). }Values are Pearson R scores. \textbf{Bold} fonts denote the best results and \underline{underlined} ones denote the second best.}}\vspace{-0.15in}
	\label{tab:splitting}
	\renewcommand{\arraystretch}{1.15}
	\begin{tabular}{@{}>{\raggedright\arraybackslash}p{2.5cm}*{7}{>{\centering\arraybackslash}p{1.1cm}}@{}}
		\toprule
		\textbf{Model}                                 & \textbf{Split} & \textbf{HLM $\uparrow$}             & \textbf{RLM$\uparrow$}             & \textbf{ER$\uparrow$}              & \textbf{Solubility$\uparrow$}          & \textbf{hPPB$\uparrow$}            & \textbf{rPPB$\uparrow$}            \\
		\midrule
		AttentiveFP~\upcite{xiong2020pushing} & Time           & \second{0.4377}          & \best{0.4694}            & 0.3783                   & 0.3917                   & 0.5843                   & 0.4222                   \\
		N-GramRF~\upcite{liu2019n}            & Time           & 0.3080                   & 0.1429                   & 0.3069                   & 0.3739                   & 0.3621                   & 0.2404                   \\
		N-GramXGB~\upcite{liu2019n}           & Time           & 0.2194                   & 0.1839                   & 0.3487                   & 0.3448                   & 0.3244                   & 0.1651                   \\
		PretrainGNN~\upcite{hu2020strategies} & Time           & 0.3357                   & 0.2222                   & 0.6413                   & \second{0.5498}          & 0.7269                   & 0.4867                   \\
		GraphMVP~\upcite{liu2021pre}          & Time           & 0.3201                   & 0.1116                   & 0.6107                   & 0.4901                   & 0.7450                   & 0.5180                   \\
		MolCLR-GCN~\upcite{wang2022molecular} & Time           & 0.1658                   & 0.1543                   & 0.6227                   & 0.3757                   & \best{0.9264}            & \best{0.7714}            \\
		MolCLR-GIN~\upcite{wang2022molecular} & Time           & 0.1452                   & 0.0938                   & 0.6192                   & 0.3556                   & 0.7726                   & 0.4187                   \\
		Mol-GDL~\upcite{shen2023molecular}    & Time           & 0.3101                   & \second{0.3065}          & \second{0.6719}          & 0.4099                   & 0.6771                   & 0.6049                   \\
		GraphKAN~\upcite{zhang2024graphkan}   & Time           & 0.1014                   & 0.1220                   & 0.6262                   & 0.3009                   & \second{0.8794}          & 0.3369                   \\
		KA-GCN~\upcite{li2025kolmogorov}     & Time           & \best{0.4596}            & 0.2324                   & 0.6204                   & \best{0.5549}            & 0.6458                   & 0.6019                   \\
		KA-GAT~\upcite{li2025kolmogorov}      & Time           & 0.2595                   & 0.2775                   & \best{0.7149}            & 0.4744                   & 0.6760                   & \second{0.6523}          \\
		\bottomrule
	\end{tabular}
\end{table*}

\section{Applications}
\label{sec:applications}
Deep learning based molecular property prediction has moved from benchmark-driven method development to deployment in end-to-end discovery pipelines. Most \rv{real-world} use cases fall into three verticals: (i) drug discovery, where property predictors act as surrogates for potency, binding affinity, ADMET, and safety, (ii) materials design, where they replace or augment quantum simulations and high throughput experiments, and (iii) broader domains such as synthetic chemistry, environmental science, and dynamics based molecular modeling. Across all three, the ``foundation model era'' is defined by pretraining on multi-million molecule or structure corpora, multi-task and multi-modal objectives spanning heterogeneous endpoints (for example, sequence, structure, text, and dynamics), and tight integration of property predictors with generative models, optimization algorithms, and simulation pipelines.

\subsection{Drug Discovery}
\label{sec:drug}
The pharmaceutical pipeline serves as a primary proving ground for molecular deep learning, where high failure costs and significant potential impact drive the field beyond relying solely on simple accuracy metrics.

\textbf{Property Prediction and ADMET Modeling.} Deep learning is integral to workflows predicting pharmacological properties and bioactivities. GNNs~\upcite{swanson_admet-ai_2024, evangelista_application_2025} and transformers~\upcite{msformer, fate-tox} model ADMET properties to prioritize molecules with favorable profiles, a capability exemplified by the data-driven discovery of the antibiotic halicin. While multitask GNNs leverage shared substructural features, they often degrade under strict \rv{scaffold-split} evaluations. Chemical foundation models~\upcite{reactiont5} address this by pretraining on extensive molecular libraries to create versatile representations. These models enable efficient few-shot prediction of pharmacokinetic properties and often match fully trained task-specific models while using significantly less data. Consequently, pipelines increasingly adopt pretrained molecular transformers and GNNs for efficient candidate triage.

\textbf{Binding Affinity}.
Protein-ligand binding affinity prediction is central to structure-based drug design. Models utilizing curated databases like PDBbind~\upcite{pdbbind} employ 3D GNN and SE(3) equivariant architectures to jointly encode protein pockets and ligands for accurate binding energy prediction. In parallel, mutation-level modeling supports affinity maturation. Datasets such as SKEMPI 2.0~\upcite{skempi2} and AB-Bind~\upcite{abbind} enable models to predict how variants alter binding free energies. This facilitates the design of higher-affinity or escape-resistant antibodies through sequence-based, structure-based, or graph-based approaches.

\textbf{Molecular Target Interaction Prediction}.
Deep learning also models molecular target interactions, evolving from classic 1D and 2D baselines~\upcite{ozturk_deepdta_2018, nguyen_graphdta_2021} to hybrid systems~\upcite{zhou2023unimol} that combine local equivariant 3D biases with global attention. A critical capability, demonstrated by T-ALPHA~\upcite{kyro2025talpha}, is achieving state-of-the-art performance using predicted rather than crystal structures. This flexibility to operate on approximate structures, such as those from AlphaFold, expands utility when high-quality co-crystallized data is unavailable.

\textbf{Generative Modeling and De Novo Design}.
Large-scale pretraining has shifted the field from screening to \emph{de novo} generation of target-aware molecules. A recent sequence-guided model~\upcite{vijil2023accelerating} discovered potent SARS-CoV-2 inhibitors with an exceptional hit rate without target-specific training. Similarly, TamGen~\upcite{tamgen} employed a GPT-like language model to design novel inhibitors for the tuberculosis ClpP protease, yielding seven experimentally validated compounds. These results illustrate how foundation models now act as creative partners to generate novel and potent chemical matter rather than merely screening existing libraries.

\subsection{Materials Design}
\label{sec:material}
Drug discovery has adapted to the foundation model paradigm, whereas many parts of materials design are being built around it. The chemical space of materials, including catalysts, electrolytes, and porous frameworks, is even more complex and combinatorial than that of small-molecule drugs. Large curated databases such as Materials Project~\upcite{materialsproj}, JARVIS~\upcite{jarvis}, and OQMD~\upcite{oqmd} provide crystal structures and computed properties at scale and serve as training corpora for deep learning models. This domain is a rich source of architectural innovation because generic models such as plain GNNs or transformers often fail and motivate hybrid and physics-informed approaches.

\textbf{Catalyst Discovery}
The design of new catalysts underpins sustainable energy, environmental remediation, and chemical manufacturing and requires accurate prediction of properties such as adsorption energies on catalyst surfaces, which were traditionally computed with expensive Density Functional Theory (DFT). DeepMind’s GNN-based model GNoME~\upcite{genome} illustrated the impact of deep learning by predicting formation energies of inorganic crystal structures at scale. Coupling GNN predictions with DFT in an active learning loop, GNoME enumerated 2.2 million candidate crystal structures and identified about 380,000 likely stable ones, an \rv{order-of-magnitude} expansion of known materials space. Hundreds of these AI-predicted crystals have been experimentally confirmed, showing that graph-based models can generalize far beyond their training distribution.

A distinctive trend in the foundation model era is an expanded notion of multimodality that includes unstructured natural language text. CatBERTa~\upcite{ock2023catberta}, a state-of-the-art model from 2023, is a transformer language model trained to predict catalyst properties such as adsorption energy directly from text descriptions of catalytic systems. It learns to associate textual descriptions extracted from the literature with corresponding DFT-calculated properties. Models such as CatBERTa~\upcite{ock2023catberta}, CataLM~\upcite{catalm}, and CatGPT~\upcite{catgpt} suggest that the scientific literature is a machine-readable data source that encodes deep chemical and physical knowledge for data-driven discovery.

\textbf{Electrolyte Formulations}
Designing energy storage materials such as liquid electrolytes for lithium-ion batteries requires modeling complex multi-component systems of salts, solvents, and additives. For these mixtures, generic GNNs are often insufficient, and \rv{state-of-the-art} performance comes from physics-informed architectures that encode fundamental physical dependencies as inductive biases. BAMBOO~\upcite{bamboo} is a representative example, a graph-equivariant and transformer-based machine learning force field that decomposes the total potential energy of organic electrolytes into semi-local, electrostatic, and dispersion contributions \rv{with explicit long-range electrostatics and dispersion, combined with an equivariant GNN backbone for short-range interactions.} 

Integrating specialized foundation models as modules enables autonomous \emph{in silico} research pipelines for next-generation batteries. The Uni-Electrolyte platform~\upcite{chen2025uni} is a complete AI-driven workflow with EMolCurator for design, EMolForger for synthesis, and EMolNetKnittor for mechanistic analysis. It demonstrates an end-to-end system that designs new candidates, checks synthetic feasibility, and explains mechanisms entirely \emph{in silico}. Similar workflows are emerging in other materials domains and are trained and evaluated on large databases such as Materials Project, JARVIS, and OQMD to cover wide compositional and structural spaces.

\subsection{Other Emerging Trends in Applications}
\label{sec:otherapp}
The impact of deep learning based molecular prediction now extends beyond pharmaceuticals and materials to synthetic chemistry, environmental science, and molecular dynamics in chemical science and engineering.

The synthesis of candidate molecules remains a bottleneck because a molecule's utility is contingent on its synthetic accessibility. Deep learning has shifted this domain from expert-coded reaction templates to data-driven sequence modeling, where reactions are framed as \rv{a translation task} between reactant and product SMILES strings. The Molecular Transformer~\upcite{schwaller2019molecular} and later foundation models such as ChemFormer~\upcite{chemformer} use large reaction corpora such as USPTO\_MIT to predict reaction outcomes and propose retrosynthetic routes with high fidelity. 

Regulatory pressure has established environmental fate and toxicity prediction as a central design constraint. Given the data scarcity for emerging pollutants, transfer learning is crucial, enabling models pretrained on pharmacological datasets to predict properties like aquatic toxicity and biodegradability. This approach is vital for PFAS screening, where conventional models often struggle with fluorinated compounds. Notably, a recent study~\upcite{Wang2025Integration} utilized a transfer learning multitask network fine-tuned on a PFAS corpus, achieving an average AUC of 0.886 across five hepatic toxicity targets, including PPAR$\alpha$ and PPAR$\gamma$.

In computational chemistry, biology, and materials science, many properties of interest are dynamical~\upcite{mlms}. Deep learning increasingly approximates potential energy surfaces and interatomic forces, enabling large-scale molecular dynamics (MD). Neural network potentials and machine learning force fields achieve near-DFT accuracy \rv{at drastically reduced costs, trained on trajectories or quantum mechanical datasets}~\upcite{mlms}. These dynamics-informed models analyze time-resolved phenomena such as phase transitions, conformational changes, transport, and reaction pathways. Moreover, applying foundation model strategies, specifically pretraining generic potentials followed by fine-tuning, effectively unifies static property prediction with dynamical behavior.~\upcite{fmmd}

Collectively, these emerging domains show how the foundation model paradigm is spreading beyond its original strongholds and how deep learning is increasingly closing the loop between static molecular representations, dynamics, and macroscopic function.

\section{Challenges and Future Directions}
\label{sec:challenges_and_future}

Molecular property prediction stands at a critical juncture, where remarkable progress in deep learning architectures is tempered by fundamental limitations in real-world applicability. This section analyzes these challenges through three lenses: architectural constraints that hinder accurate modeling of molecular systems, evaluation practices that obscure true model readiness, and translational gaps that prevent laboratory adoption. We then present a research agenda grounded in physics-aware learning, robust benchmarking, and interdisciplinary co-development.

\subsection{Architectural and Algorithmic Challenges}
\label{sec:arch_challenges}
Current deep learning approaches face significant limitations in handling molecular complexity. \rv{Graph neural networks, while effective for local interactions, struggle to capture long-range quantum effects critical for properties like protein-ligand binding affinity, as reflected in their limited performance on binding affinity benchmarks such as PDBbind~\upcite{wang2005pdbbind} compared to well-calibrated physics-based scoring functions.} Transformer architectures address some of these limitations through global attention mechanisms but suffer from quadratic computational scaling that renders them impractical for macromolecular systems. Even state-of-the-art geometric models like SE(3)-equivariant networks~\upcite{fuchs2020se} remain computationally prohibitive for systems beyond 100 atoms, \rv{while still learning approximate mappings rather than encoding exact quantum mechanical relationships}.

The field also grapples with a transparency crisis. High-performing models such as MolBERT~\upcite{li2021mol} achieve impressive benchmark results but operate as black boxes, providing little insight into the physicochemical basis of their predictions. This interpretability deficit creates barriers for regulatory adoption in drug discovery, where mechanistic understanding is \rv{essential}. \rv{Additionally, efforts to incorporate multimodal information by integrating structural graphs with spectral data or knowledge from literature have yet to demonstrate consistent improvements, pointing to underlying limitations in existing data fusion techniques.}

\subsection{Evaluation and Benchmarking Shortcomings}
\label{sec:eval_shortcomings}
\rv{Building on the dataset limitations and evaluation protocol guidance in Section~\ref{sec:benchmarks}, we summarize recurring shortcomings in MPP benchmarking that can inflate perceived progress and hinder reproducibility~\upcite{Ash2025Protocols}.}

\begin{itemize}
    \item \rv{\textbf{Single-split, single-seed reporting.} Many studies report a single score (often from one split and one seed) and treat small differences as definitive. In practice, rankings can be unstable due to data scarcity, label noise, class imbalance, and training stochasticity. Robust comparisons should quantify variability via repeated evaluation and report distributions (e.g., mean$\pm$std or confidence intervals)~\upcite{Ash2025Protocols}.}
    
    \item \rv{\textbf{Split-induced leakage and overly optimistic generalization.} Random splits can overestimate performance when chemically similar compounds appear in both train and test sets. New splitting methods (e.g., Butina, UMAP) reduce this leakage and often better probe extrapolation to new chemotypes, but conclusions must still be scoped to the evaluated split regime~\upcite{bemis1996properties,sheridan2004similarity,walters2024splitting}.}
    
    \item \rv{\textbf{Insufficient statistical inference.} Even with repeated runs, many comparisons omit statistical tests or apply tests that ignore correlation across folds/repeats and multiple-comparison issues. Non-parametric multi-dataset testing with post-hoc analysis is often more appropriate, and corrected tests (or Bayesian alternatives) can improve interpretability under repeated CV~\upcite{demsar2006statistical,dietterich1998approximate}.}
    
    \item \rv{\textbf{Metric mismatch with decision-making.} Default metrics (e.g., ROC-AUC) may not reflect operational goals (e.g., early enrichment, low-FPR precision) and regression gains should be interpreted relative to experimental noise. Recent guidance emphasizes selecting decision-relevant metrics and reporting \emph{practical} significance (effect sizes / meaningful margins), not only statistical significance~\upcite{Ash2025Protocols}.}
\end{itemize}

\srv{A further limitation is that most current MPP benchmarks remain narrowly unimodal. They typically evaluate prediction from molecular structures alone, while giving limited attention to how models handle other scientifically relevant modalities such as spectra, assay metadata, procedural text, structural images, or other structured experimental context. Recent multimodal chemistry benchmarks reinforce this concern. In USNCO-V, Cui et al.~\upcite{cui2025multimodal} showed that for some weaker models, removing image inputs could slightly improve accuracy, indicating that visual information may introduce noise when cross-modal alignment is poor. In MaCBench, Alampara et al.~\upcite{alampara2025probing} found that when equivalent scientific information was presented in textual rather than visual form, models typically performed better on the text version, highlighting persistent limitations in modality fusion and cross-modal scientific reasoning. Although these benchmarks are not property-prediction benchmarks per se, they expose a broader evaluation gap that is highly relevant to molecular AI.}
 
\srv{Collectively, we encourage benchmark studies to (i) specify the split protocol and all randomness sources, (ii) report uncertainty over repeated evaluations, (iii) apply statistically sound comparisons across tasks and datasets, (iv) prioritize effect sizes and decision-aligned metrics so that improvements are both reproducible and actionable, and (v) gradually expand benchmark design toward richer, modality-aware evaluation settings that better reflect the heterogeneous evidence streams used in real chemical research~\upcite{Ash2025Protocols,demsar2006statistical,alampara2025probing,cui2025multimodal}.}

\subsection{Practical Adoption Barriers}
\label{sec:barriers}
\srv{The gap between computational predictions and experimental validation represents one of the most pressing challenges. Few models undergo rigorous wet-lab validation, and those that do often show significantly degraded performance in real-world settings~\upcite{Fang2023Prospective}. This translational gap stems from multiple factors: inadequate uncertainty quantification that fails to distinguish reliable predictions from speculative ones; oversimplification of molecular representations that ignore crucial solvent or temperature effects; \sxsrv{limited robustness when models are extended from relatively simple perception tasks to workflows requiring multi-step reasoning over textual and spatial inputs}; and a persistent disconnect between machine learning researchers and experimental chemists in problem formulation.}

\srv{As discussed in Section~\ref{sec:uq}, a range of UQ methodologies have been developed for molecular property prediction, from ensemble-based and mean-variance estimation approaches to evidential regression and conformal prediction. In the context of MPP deployment, practical pipelines such as Chemprop~\upcite{heid_chemprop_2024} integrate multiple methods to balance accuracy, uncertainty quality, and computational feasibility. Uncertainty calibration has also been explored at the output level: Schwaller et al.~\upcite{schwaller2019molecular} demonstrated calibrated confidence estimates in a Molecular Transformer for reaction prediction, and Kyro et al.~\upcite{kyro2025talpha} incorporated uncertainty-aware self-learning into protein-ligand binding affinity prediction.}

\srv{Despite this progress, current UQ methods still face major barriers to laboratory adoption. As discussed in Section~\ref{sec:arch_challenges}, architectural limitations such as the finite receptive field of message-passing networks~\upcite{MPNN} can impair the modeling of long-range interactions and thus degrade property prediction itself. These limitations also affect uncertainty estimation: when the underlying model cannot represent the relevant physics, uncertainty scores may inherit the same systematic blind spots~\upcite{gasteiger2020fast}. In addition, benchmark studies show that UQ methods perform inconsistently across datasets, with no universally superior approach~\upcite{hirschfeld2020uncertainty, dai2025uncertainty}. Computationally intensive methods such as deep ensembles are often difficult to scale, while more efficient single-pass methods may require careful recalibration.}

\srv{These issues become even more important in the foundation-model regime. Foundation models are especially attractive for low-label tasks and out-of-distribution settings, but these are also the settings in which reliability is hard to assess. Strong average transfer performance does not guarantee that a specific molecule falls within the model's effective applicability domain, nor that the pretrained representation captures the relevant chemotype, assay context, or structure-property relationship. Calibrated uncertainty is therefore essential for distinguishing reliable interpolation from risky extrapolation, prioritizing experiments, and enabling risk-aware deployment. The next step is thus not merely larger molecular foundation models, but foundation models with scalable and well-calibrated uncertainty estimation.}

\srv{\sxsrv{A related barrier arises in multimodal and foundation-model workflows that must reason jointly over molecular strings, structures, and spatial information. Models that perform well on comparatively simple perception-style tasks may generalize poorly when extended to analytical settings requiring multi-step reasoning. Cui et al.~\upcite{cui2025multimodal} showed that fragile modality integration can prevent richer inputs from improving predictions and, in some cases, can even reduce analytical accuracy. Alampara et al.~\upcite{alampara2025probing} further reported persistent weaknesses in spatial reasoning and multi-step logical inference, including settings involving isomers and crystal space groups, with performance degrading sharply as the reasoning chain length increases. These findings suggest that multimodal scaling alone is insufficient unless models can reliably align heterogeneous inputs and sustain coherent reasoning across them.
}}

\srv{Beyond barriers mentioned above, adoption is further limited by oversimplified representations and insufficient interdisciplinary coordination during model development and validation~\upcite{Ash2025Protocols, Fang2023Prospective}. These challenges highlight the need for benchmarks that better reflect real-world complexity and support the development of uncertainty-aware, experimentally validated MPP systems.}

\subsection{Pathways to Advancement}
\label{sec:pathway}
Three interconnected research directions emerge as critical for overcoming current limitations, corresponding to the roadmap outlined in Figure~\ref{fig:intro}.

\textbf{Physics-aware learning.} As discussed in Sections~\ref{sec:gnn}, \ref{sec:quantum}, and~\ref{sec:arch_challenges}, future progress depends on models that better align predictive performance with molecular geometry and physical mechanism. Directional and geometry-aware message passing networks~\upcite{dimenet,gasteiger2020fast,yuan2023molecular} have shown that embedding angular and spatial priors can substantially improve geometry-sensitive prediction. More broadly, combining symmetry-aware representations with physically constrained objectives offers a principled path toward more reliable generalization across chemical space.

\textbf{Uncertainty-calibrated foundation models.} As discussed in Sections~\ref{sec:uq} and~\ref{sec:barriers}, foundation models are especially valuable in low-label and out-of-distribution settings, but these are also the settings in which reliability is hardest to assess. Strong average transfer performance alone is therefore insufficient. Future models should couple large-scale pretraining with scalable, well-calibrated uncertainty estimation that distinguishes reliable interpolation from risky extrapolation, identifies when molecules fall outside the model's effective applicability domain, and supports experimental prioritization and risk-aware deployment.

\textbf{Realistic multimodal benchmark ecosystems.} As detailed in Sections~\ref{sec:benchmarks} and~\ref{sec:eval_shortcomings}, current benchmarks remain largely structure-centric and are affected by data curation issues, split-induced leakage, metric inflation, and inconsistent evaluation practices. Future benchmark design should therefore incorporate richer evidence streams, such as spectra, assay metadata, procedural information, and structural images, while adopting more rigorous, deployment-realistic protocols that better reflect real scientific use cases~\upcite{Ash2025Protocols}. In parallel, multimodal models should be evaluated not only by aggregate accuracy, but also by their ability to align heterogeneous inputs, remain robust to noisy modalities, and support multi-step scientific reasoning beyond superficial pattern matching~\upcite{cui2025multimodal,alampara2025probing}.

Beyond these three core directions, complementary themes such as improving data efficiency through self-supervised pretraining~\upcite{fang2022geometry,mole,chemberta} and active learning~\upcite{fund_model_rev}, embedding domain knowledge into model architectures~\upcite{Alhmoudi2025,mole}, and developing translational tools to bridge the computational--experimental divide also merit attention, though they are not explored in depth here.

In summary, advancing the field requires physically grounded learning, uncertainty-calibrated foundation models, and realistic multimodal benchmarks that better reflect real chemical research. Together, these priorities can help MPP evolve from benchmark-driven modeling toward more trustworthy and practically useful molecular intelligence.

\section{Conclusion}
\label{sec:conclu}

\rv{Molecular property prediction has evolved through four complementary methodological eras, from first-principles quantum mechanics to data-driven foundation models, each contributing unique advantages that remain valuable across different problem scales. This cumulative progression has enabled remarkable accuracy on standard benchmarks while exposing persistent challenges in generalization, scalability, and uncertainty quantification. Our unified taxonomy (Table~\ref{tab:unified_framework}) reveals how these paradigms collectively structure the field: quantum methods provide physically interpretable baselines; descriptor-based approaches remain efficient for low-data regimes; geometric deep learning architectures excel at capturing 3D symmetries; and foundation models integrate multimodal, cross-domain knowledge. Together, these methods form a complementary toolkit rather than a replacement hierarchy.}

\rv{Looking ahead, progress in molecular AI will likely depend on three convergent directions: (1) \textit{physics-aware learning} that enforces consistency with quantum principles; (2) \textit{uncertainty-calibrated foundation models} trained on large, diverse, and multimodal corpora; and (3) \textit{realistic multimodal benchmarks} that reflect the structural and chemical heterogeneity of real-world systems. Together, these directions may enable the development of general-purpose molecular modeling frameworks capable of accurately simulating chemical behavior and guiding molecular design across quantum, biological, and materials domains. Such frameworks could help bridge computation and experiment, accelerating discovery in drug design, sustainable materials, and beyond.}

\bibliographystyle{unsrt} 
\bibliography{ref}
\appendix

\section{Standard Loss Functions for Learning Paradigms}
\label{app:losses}

This appendix provides the formal loss function definitions for the learning paradigms in Section~\ref{sec:learning-settings}.

\textbf{Supervised learning.}
Given a labeled dataset $\mathcal{D}_L = \{(m_i, y_i)\}_{i=1}^N$, the model $f_\theta$ is trained by minimizing a task-specific loss:
\begin{equation}
	\min_\theta \frac{1}{N}\sum_{i=1}^N \mathcal{L}_{\mathrm{task}}\big(f_\theta(m_i),\, y_i\big).
\end{equation}
For regression tasks with continuous targets $y\in\mathbb{R}$, the loss is typically the mean squared error:
\begin{equation}
	\mathcal{L}_{\mathrm{MSE}} = \frac{1}{N} \sum_{i=1}^N \big(f(m_i) - y_i\big)^2.
\end{equation}
For classification tasks with discrete labels $y \in \{1, \dots, C\}$, cross-entropy is used:
\begin{equation}
	\mathcal{L}_{\mathrm{CE}} = -\sum_{c=1}^C y_c \log f(m)_c.
\end{equation}

\textbf{Self-supervised pretraining.}
Given an unlabeled corpus $\mathcal{D}_U = \{m_i\}_{i=1}^M$, self-supervised learning minimizes a proxy loss:
\begin{equation}
	\min_\theta \sum_{m \in \mathcal{D}_U} \mathcal{L}_{\mathrm{SSL}}(m, f_\theta).
\end{equation}
Common instantiations include contrastive objectives that maximize agreement between augmented views $m', m''$ of the same molecule via cosine similarity $\mathrm{sim}(\cdot,\cdot)$, and reconstructive objectives that train the model to recover masked or corrupted inputs $\tilde{m}$ by minimizing $\mathcal{L}_{\mathrm{recon}}(f_\theta(\tilde{m}), m)$.

\textbf{Transfer learning and fine-tuning.}
Starting from pretrained parameters $\theta_{\mathrm{pre}}$, fine-tuning optimizes:
\begin{equation}
	\min_\theta \sum_{(m, y) \in \mathcal{D}_L} \mathcal{L}_{\mathrm{task}}\big(f_\theta(m),\, y\big), \quad \theta \leftarrow \theta_{\mathrm{pre}}.
\end{equation}
Multi-task variants jointly minimize $\sum_{k=1}^K \lambda_k \mathcal{L}_{\mathrm{task}}^{(k)}$ over $K$ related tasks with task-specific weights $\lambda_k$.

\textbf{Semi-supervised learning.}
A combined loss integrates supervised and unsupervised components:
\begin{equation}
	\mathcal{L}_{\mathrm{total}} = \mathcal{L}_{\mathrm{supervised}} + \lambda\, \mathcal{L}_{\mathrm{unsupervised}},
\end{equation}
where $\lambda$ controls the relative contribution of the unsupervised regularizer.

\section{Formal Symmetry Definitions for 3D Molecular Representations}
\label{app:symmetry}

This appendix provides the formal mathematical treatment of geometric encodings and spatial symmetry constraints. These definitions underpin the 3D representation modalities discussed in Section~\ref{sec:3d-rep} and the geometric architectures of Section~\ref{sec:architectures}.

\subsection{Geometric Encoding Formulations}

Let a molecule be $\mathcal{M}=\{(Z_i,\mathbf{r}_i)\}_{i=1}^N$ with $\mathbf{r}_i\in\mathbb{R}^3$ ($N$: number of atoms; $Z_i$: atomic number; $\mathbf{r}_i$: Cartesian coordinates).

\textbf{Distance-based features.}
Pairwise interatomic distances are defined as $d_{ij}=\|\mathbf{r}_j-\mathbf{r}_i\|$. These scalar distances are typically expanded through radial basis functions to produce continuous, high-dimensional representations:
\begin{equation}
	\mathrm{RBF}_m(d_{ij})=\exp\!\left(-\frac{(d_{ij}-\mu_m)^2}{2\sigma^2}\right),
\end{equation}
with learnable or fixed centers $\mu_m$ and width $\sigma>0$.

\textbf{Directional encodings.}
Bond angles between atom triplets $(i,j,k)$ and dihedral angles between atom quadruplets $(i,j,k,l)$ are defined as:
\begin{equation}
	\cos\theta_{ijk}=\frac{(\mathbf{r}_i-\mathbf{r}_j)\cdot(\mathbf{r}_k-\mathbf{r}_j)}{\|\mathbf{r}_i-\mathbf{r}_j\|\,\|\mathbf{r}_k-\mathbf{r}_j\|},
\end{equation}
\begin{equation}
	\phi_{ijkl}=\arctan\!\frac{\hat{\mathbf{r}}_{jk}\cdot(\mathbf{n}_1\times\mathbf{n}_2)}{\mathbf{n}_1\cdot\mathbf{n}_2},
\end{equation}
where $\mathbf{n}_1=(\mathbf{r}_i-\mathbf{r}_j)\times(\mathbf{r}_k-\mathbf{r}_j)$ and $\mathbf{n}_2=(\mathbf{r}_j-\mathbf{r}_k)\times(\mathbf{r}_l-\mathbf{r}_k)$ are the normal vectors of adjacent planes.

\textbf{Spherical-basis encodings.}
Each neighbor pair is represented via a product of radial and angular bases:
\begin{equation}
	\psi_{n\ell m}^{(ij)} = g_n(d_{ij})\,Y_\ell^m(\theta_{ij},\varphi_{ij}),
\end{equation}
where $(d_{ij},\theta_{ij},\varphi_{ij})$ are the spherical coordinates of $\mathbf{r}_{ij}=\mathbf{r}_j-\mathbf{r}_i$, $g_n$ is a radial basis (e.g., Bessel or raised-cosine), and $Y_\ell^m$ are real spherical harmonics of degree $\ell$ and order $m$. The indices $n,\ell,m$ control expressiveness and computational cost~\upcite{spherenet}.

\subsection{Spatial Symmetry Constraints}
\label{app:symmetry-constraints}

Molecular properties that are scalar quantities (e.g., energy, solubility) must be invariant to rigid motions of $\mathbb{R}^3$. Formally, with $\mathbf{Q}\in\mathrm{O}(3)$ (rotation/reflection) and $\mathbf{t}\in\mathbb{R}^3$ (translation):

\textbf{E(3) invariance.}
A prediction function $f$ is E(3)-invariant if
\begin{equation}
	f\big(\{\mathbf{Q}\mathbf{r}_i+\mathbf{t}\}\big) = f\big(\{\mathbf{r}_i\}\big), \quad \forall\, \mathbf{Q}\in\mathrm{O}(3),\;\mathbf{t}\in\mathbb{R}^3.
\end{equation}
This is typically enforced by building features solely from geometric invariants $\{d_{ij},\theta_{ijk},\phi_{ijkl},\dots\}$.

\textbf{E(3) equivariance.}
Equivariant models maintain intermediate features $\Phi$ that transform predictably under spatial operations:
\begin{equation}
	\Phi\big(\{\mathbf{Q}\mathbf{r}_i+\mathbf{t}\}\big) = \rho(\mathbf{Q})\,\Phi\big(\{\mathbf{r}_i\}\big),
\end{equation}
where $\rho$ is a representation of the group in feature space. For vector features, $\rho(\mathbf{Q})=\mathbf{Q}$; for higher-order tensor features, $\rho$ acts via Wigner-$D$ matrices $\mathbf{D}^{(\ell)}(\mathbf{Q})$. The final scalar prediction is obtained by a permutation- and rotation-invariant readout, e.g., $y=\mathrm{pool}_i\,\Phi_i^{(\ell=0)}$ (pooling over scalar components only).

\textbf{Equivariant coordinate updates.}
Some architectures~\upcite{EGNN} additionally update atomic positions in an equivariant manner:
\begin{equation}
	\mathbf{r}_i'=\mathbf{r}_i+\sum_{j\in\mathcal{N}(i)}\beta_{ij}\,\hat{\mathbf{r}}_{ij}, \quad \hat{\mathbf{r}}_{ij} = \frac{\mathbf{r}_j-\mathbf{r}_i}{\|\mathbf{r}_j-\mathbf{r}_i\|},
\end{equation}
where the scalar weights $\beta_{ij} = \phi_{\beta}(\mathbf{h}_i,\mathbf{h}_j,d_{ij})$ depend only on rotationally invariant inputs. This ensures that the updated coordinates $\{\mathbf{r}_i'\}$ transform equivariantly under E(3), while scalar property predictions obtained from invariant readouts remain invariant.

\section{Hyperparameter Settings}
\label{app:hyperparams}

Table~\ref{tab:hyperparameters} presents hyperparameter configurations for models in experiments. To ensure fair comparison across different architectures, we employ a unified molecular encoding scheme using the CGCNN atom encoder and a 14-dimensional bond encoder for all models. The key hyperparameters include the number of GNN layers, hidden dimension size, dropout rate, pooling strategy, learning rate, number of training epochs, and batch size. Most models are trained with 2-5 layers, learning rates ranging from $10^{-4}$ to $10^{-3}$, and a consistent batch size of 128 across 501 training epochs. Dropout rates vary from 0.1 to 0.5 depending on the model architecture. Some specialized models incorporate additional parameters, such as grid size and spline order for KA-GNN architectures, or attention heads for graph attention variants.

\begin{table}[H]
	\centering
	\scriptsize
	\caption{Hyperparameter settings for all evaluated models. A dash (–) indicates the parameter uses default settings or is not applicable.}
	\label{tab:hyperparameters}
	\vspace{-0.1in}
	\resizebox{\textwidth}{!}{
		\begin{tabular}{lcccccccl}
			\toprule[1.2pt]
			\rowcolor{white!10}
			\textbf{Model} & \textbf{Layers} & \textbf{Hidden} & \textbf{Dropout} & \textbf{Pooling} & \textbf{LR} & \textbf{Epochs} & \textbf{Batch} & \textbf{Special} \\
			\midrule[0.8pt]
			AttentiveFP    & 2               & --              & --               & MEAN/AVG      & 1e-4        & 501             & 128            & --               \\
			N-GramRF       & 3               & --              & --               & MEAN/AVG      & 1e-3        & 501             & 128            & --               \\
			N-GramXGB      & 3               & --              & --               & MEAN/AVG      & 1e-3        & 501             & 128            & --               \\
			PretrainGNN    & 5               & --              & 0.5              & MEAN/AVG      & 1e-3        & 501             & 128            & --               \\
			GraphMVP       & 5               & --              & 0.5              & MEAN/AVG      & 1e-3        & 501             & 128            & --               \\
			MolCLR-GCN     & 5               & 256             & 0.3              & MEAN/AVG      & 5e-4        & 501             & 128            & --               \\
			MolCLR-GIN     & 5               & 512             & 0.3              & MEAN/AVG      & 5e-4        & 501             & 128            & --               \\
			Mol-GDL        & 3               & --              & --               & MEAN/AVG      & 1e-4        & 501             & 128            & --               \\
			GraphKAN       & 3               & 256             & 0.1              & MEAN/AVG      & 1e-3        & 501             & 128            & grid=5, order=3  \\
			KA-GCN         & 4               & --              & --               & MEAN/AVG      & 1e-4        & 501             & 128            & --               \\
			KA-GAT         & 2               & --              & --               & MEAN/AVG      & 1e-4        & 501             & 128            & grid=3, heads=2  \\
			\bottomrule[1.2pt]
		\end{tabular}
	}
	\vspace{-0.05in}
\end{table}


\clearpage
\end{document}